\documentclass[10pt,twocolumn,letterpaper]{article}

\usepackage{iccv}
\usepackage{times}
\usepackage{epsfig}
\usepackage{graphicx}
\usepackage{amsmath}
\usepackage{amssymb}
\usepackage{textcomp}

\usepackage{eqnarray}
\usepackage{amsmath}
\usepackage{units}
\usepackage{graphicx}
\usepackage{caption}
\usepackage{subcaption}
\usepackage{url}
\usepackage{algorithmicx}
\usepackage{algpseudocode}
\usepackage{mathtools}
\usepackage{bbm}
\usepackage{stackengine}
\usepackage{xfrac}

\usepackage{multirow}
\usepackage{colortbl}

\usepackage[breaklinks=true,bookmarks=false]{hyperref}

\iccvfinalcopy % *** Uncomment this line for the final submission

\def\iccvPaperID{552} % *** Enter the ICCV Paper ID here

\DeclarePairedDelimiter{\norm}{\lVert}{\rVert}

\usepackage{expl3}
\ExplSyntaxOn
\newcommand\latinabbrev[1]{
  \peek_meaning:NTF . {% Same as \@ifnextchar
    #1\@}%
  { \peek_catcode:NTF a {% Check whether next char has same catcode as \'a, i.e., is a letter
      #1.\@ }%
    {#1.\@}}}
\ExplSyntaxOff

\def\etal{\latinabbrev{et al}}

\definecolor{Yellow}{rgb}{1,1, 0.6}
\definecolor{Red}{rgb}{1, 0.6, 0.6}

\ificcvfinal\fi
\begin{document}

%%%%%%%%% TITLE

\title{Convolutional Color Constancy}

\author{Jonathan T. Barron\\
{\tt\small barron@google.com}
}

\maketitle

\begin{abstract}
Color constancy is the problem of inferring the color of the light that illuminated a scene, usually so that the illumination color can be removed.
Because this problem is underconstrained, it is often solved by modeling the statistical regularities of the colors of natural objects and illumination.
In contrast, in this paper we reformulate the problem of color constancy as a 2D spatial localization task in a log-chrominance space, thereby allowing us to apply techniques from object detection and structured prediction to the color constancy problem.
By directly learning how to discriminate between correctly white-balanced images and poorly white-balanced images, our model is able to improve performance on standard benchmarks by nearly $40\%$.
\end{abstract}

\section{Intro}

\begin{figure}[b!]
\centering
    \stackunder[10pt]{
    \stackunder[5pt]{
      \includegraphics[width=1.25in]{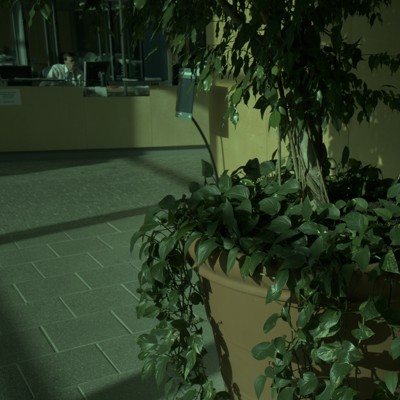}
      }{$I$}
    \stackunder[5pt]{
      \hspace{0.0in}
      }{$=$}
    \stackunder[5pt]{
      \includegraphics[width=1.25in]{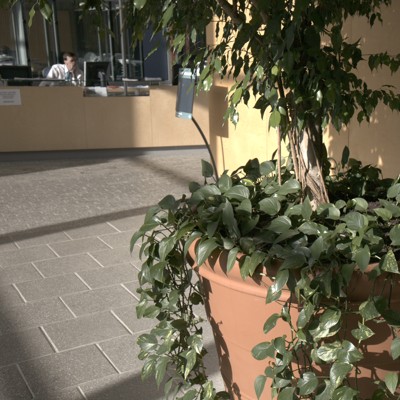}
    }{$W$}
    \stackunder[5pt]{
      \hspace{0.0in}
    }{$\times$}
    \stackunder[5pt]{
      \includegraphics[width=0.156in]{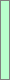}
      }{$L$}
    }{
    \stackunder[5pt]{
    \includegraphics[width=1.25in]{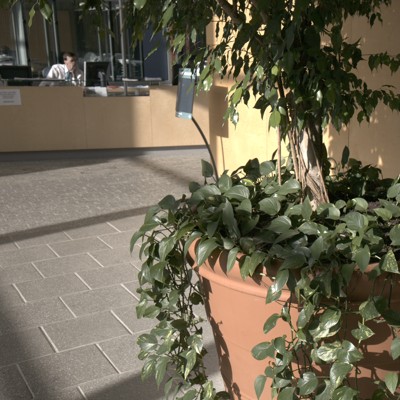}
    \hspace{-0.05in}
     \includegraphics[width=0.156in]{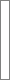}
    }
    { our $\hat W$, $\hat L$, err = $0.13$\textdegree}
    \hspace{0.13in}
    \stackunder[5pt]{
    \includegraphics[width=1.25in]{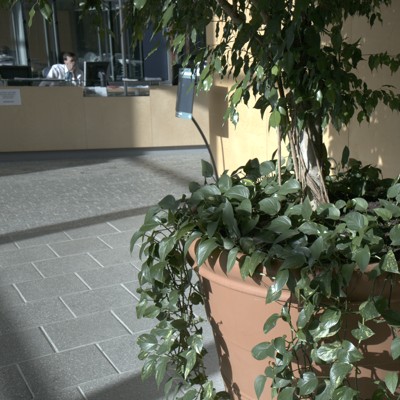}
    \hspace{-0.05in}
     \includegraphics[width=0.156in]{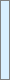}
     }
    { baseline $\hat W$, $\hat L$, err = $5.34$\textdegree}
    }

    \caption{Here we demonstrate the color constancy problem: the input image $I$ (taken from \cite{Gehler08,shifunt}) looks green, and we want to recover a white-balanced image $W$ and illumination $L$ which reproduces $I$.
    Below we have our model's solution and error for this image compared to a state of the art baseline \cite{Finlayson2013} (recovered illuminations are rendered with respect to ground-truth, so white is correct).
    More results can be seen in the supplement.
    \label{fig:problem}
}
\end{figure}

The color of a pixel in an image can be described as a product of two quantities: reflectance (the color of the paint of the surfaces in the scene) and illumination (the color of the light striking the surfaces in the scene).
When a person stands in a room lit by a colorful light they unconsciously ``discount the illuminant'', in the words of Helmholtz \cite{Gilchrist06}, and perceive the objects in the room as though they were illuminated by a neutral, white light.
Endowing a computer with the same ability is difficult, as this problem is fundamentally underconstrained --- given a yellow pixel, how can one discern if it is a white object under a yellow illuminant, or a yellow object under a white illuminant? The most general characterization of this problem is the ``intrinsic image'' problem \cite{BarrowTenenbaum78}, but the specific problem of inferring and correcting the color of the illumination of an image is commonly referred to as ``color constancy'' or ``white balance''.
A visualization of this problem can be seen in Figure~\ref{fig:problem}.

Color constancy is a well studied in both vision science and computer vision, as it relates to the academic study of human perception as well as practical problems such as designing an object recognition algorithm or a camera.
Nearly all algorithms for this task work by assuming some regularity in the colors of natural objects viewed under a white light.
The simplest such algorihm is ``gray world'', which assumes that the illuminant color is the average color of all image pixels, thereby implicitly assuming that object reflectances are, on average, gray \cite{Buchsbaum80}.
This simple idea can be generalized to modeling gradient information or using generalized norms instead of a simple arithmetic mean \cite{Barnard02, vandeWeijerTIP2007}, modeling the spatial distribution of colors with a filter bank \cite{Chakrabarti11}, modeling the distribution of color histograms \cite{FinlaysonTPAMI2001}, or implicitly reasoning about the moments of colors using PCA \cite{Cheng14}.
Other models assume that the colors of natural objects lie with some gamut \cite{ Barnard00, Forsyth1990}.
Most of these models can be thought of as statistical, as they either assume some distribution of colors or they learn some distribution of colors from training data.
This connection to learning and statistics is sometimes made more explicit, often in a Bayesian framework \cite{Brainard97bayesiancolor, Gehler08}.

One thing that these statistical or learning-based models have in common is that they are all \emph{generative} models of natural colors --- a model is learned from (or assumed of) white-balanced images, and then that model is used to white-balance new images.
In this paper, we will operate under the assumption that white-balancing is a \emph{discriminative} task.
That is, instead of training a generative model to assign high likelihoods to white-balanced images under the assumption that such a model will perform well at white-balancing, we will explicitly train a model to distinguish between white-balanced images and non-white-balanced images.
This use of discriminative machine learning appears to be largely unexplored in context of color constancy, though similar tools have been used to augment generative color constancy models with face detection \cite{Bianco12} or scene classification \cite{GijsenijTPAMI2011} information.
The most related technique to our own is probably that of Finlayson \cite{Finlayson2013} in which a simple ``correction'' to a generalized gray-world algorithm is learned using iterative least-squares, producing state-of-the-art results compared to prior art.

Let us contrast the study of color constancy algorithms with the seemingly disparate problem of object detection.
Object detection has seen a tremendous amount of growth and success in the last 20 years owing in large part to standardized challenges \cite{Everingham2010} and effective machine learning techniques, with techniques evolving from simple sliding window classifiers \cite{Dalal2005,Rowley1998,Viola2001} to sophisticated deformable models \cite{felzenszwalb2010} or segmentation-based techniques \cite{girshick2014}.
The vast majority of this work operates under the assumption that object detection should be phrased as the problem of learning a discriminative classifier which predicts whether an image patch is, in the case of face detection for example, a ``face'' or a ``nonface''.
It is common knowledge that reasoning about the ``nonface'' background class is nearly as important as reasoning about the object category of interest, as evidenced by the importance of ``mining for hard negatives'' \cite{Rowley1998} when learning an effective object detection system.
In contrast, training a generative model of an object category for detection is widely considered to be ineffective, with some unusual exceptions \cite{BharathECCV2012}.
At first glance, it may seem that all of this has little to teach us about color constancy, as most established color constancy algorithms are fundamentally incompatible with the discriminative learning techniques used in object detection.
But if the color constancy problem could be reduced to the problem of localizing a template in some $n$-dimensional space, then presumably the lessons learned from object detection techniques could be used to design an effective color constancy algorithm.

In this paper we present CCC (``Convolutional Color Constancy''), a novel color constancy algorithm that has been designed under the assumption that color constancy is a discriminative learning task.
Our algorithm is based around the observation that scaling the color channels of an image induces a translation in the log-chromaticity histogram of that image.
This observation allows us to frame the color constancy problem as a discriminative learning problem, using tools similar to convolutional neural networks \cite{lecun1989backpropagation} and structured prediction \cite{taskar2005}.
Effectively, we are able to reframe the problem of color constancy as the problem of localizing a template in some two-dimensional space, thereby allowing us to borrow techniques from the well-understood problem of object detection.
By discriminatively training a color constancy algorithm in this way, we produce state-of-the-art results and reduce error rates on standard benchmarks by nearly $40\%$.

Our paper will proceed as follows: In Section~\ref{sec:overview} we will demonstrate the relationship between image tinting and log-chrominance translation.
In Section~\ref{sec:problem} we will describe how to learn a discriminative color constancy algorithm in our newly-defined log-chrominance space.
In Section~\ref{sec:filtering} we will explain how to perform efficient filtering in our log-chrominance space, which is required for fast training and evaluation.
In Section~\ref{sec:channels} we will show how to generalize our model from individual pixel colors to spatial phenomena like edges and patches.
In Section~\ref{sec:results} we will evaluate our model on two different color constancy tasks, and in Section~\ref{sec:conclusion} we will conclude.

\section{Image Formation}
\label{sec:overview}

Consider a photometric linear image $I$ taken from a camera, in which black-level correction has been performed and in which no pixel values have saturated.
According to a simplified model of image formation, each RGB pixel value in $I$ is the product of the ``true'' white-balanced RGB value $W$ for that pixel and the RGB illumination $L$ shared by all pixels, as shown in Figure~\ref{fig:problem}.
\begin{align}
I = W \times L
\label{eq:image_formation_rgb}
\end{align}
This is a severe oversimplification of the larger ``intrinsic image'' problem, as it ignores shading, reflectance properties, spatially-varying illumination, etc.
This model also assumes that color constancy can be achieved by simply modifying the gains of each channel individually (the Von Kries coefficient law \cite{vonKies1905}) which, though certainly an approximation \cite{Brainard86}, is an effective and widespread assumption.
Our goal is, given $I$, to estimate $L$ and then produce $W = I / L$.

Let us define two measures of chrominance $u$ and $v$ from the RGB values of $I$ and $W$:
\begin{align}
I_{u} &= \log(I_g / I_r) & I_{v} &= \log(I_g / I_b) \nonumber \\
W_{u} &= \log(W_g / W_r) & W_{v} &= \log(W_g / W_b)
\end{align}
Additionally, it is convenient to define a luminance measure $y$ for $I$:
\begin{equation}
I_y = \sqrt{I_r^2 + I_g^2 + I_b^2}%\min\left(I_r, I_g, I_b \right)
\end{equation}
%This quantity indicates how confident we can be in $u$ and $v$ --- if any of $r$, $g$, or $b$ are near zero than than at least one chrominance measure is likely noisy or incorrect.
Given that we do not care about the absolute scaling of $W$, the problem of estimating $L$ simplifies further to just estimating the ``chrominance'' of $L$, which can just be represented as two numbers:
\begin{align}
L_u &= \log(L_g / L_r) & L_v &= \log(L_g / L_b)
\end{align}
Notice that by our definitions and by the properties of logarithms, we can rewrite the problem formulation in Equation~\ref{eq:image_formation_rgb} in this log-chrominance space:
\begin{align}
W_{u} &= I_{u} - L_{u} & W_{v} &= I_{v} - L_{v}
\label{eq:image_formation_uv}
\end{align}
So, our problem reduces to recovering just two quantities: $(L_{u}, L_{v})$.
Because of the absolute scale ambiguity, the inverse mapping from RGB to UV is undefined.
So after recovering $(L_u, L_v)$, we make the additional assumption that $L$ is unit-norm which allows us to recover $(L_r, L_g, L_b)$:
\begin{align}
L_r &= {\exp(-L_u) \over z} \qquad L_g = {1 \over z} \qquad L_b = {\exp(-L_v) \over z} \nonumber \\
z &= \sqrt{\exp(-L_u)^2 + \exp(-L_v)^2 + 1}
\end{align}
This log-chrominance formulation has several advantages over the RGB formulation.
We have 2 unknowns instead of 3, and we just have a simple linear constraint relating $W$ and $I$ instead of a multiplicative constraint.
Though they may seem unimportant, these properties are required to reformulate our problem as a 2D spatial localization task.

\section{Learning}
\label{sec:problem}

Let us consider an input image $I$ and its ground-truth illumination $L$.
We will construct a histogram $M$ from $I$, where $M(u,v)$ is the the number of pixels in $I$ whose chrominance is near $(u, v)$, with histogram counts weighted by each pixel's luminance:
\begin{equation}
\resizebox{2.9in}{!}{$
M(u,v) = \displaystyle \sum_i I_y^{(i)} \left[ \left|I_u^{(i)} - u \right| \leq {\epsilon \over 2} \wedge  \left| I_v^{(i)} - v \right| \leq {\epsilon \over 2} \right] $}
\end{equation}
Where the square brackets are an indicator function and $\epsilon$ is the bin-width of the histogram (in all experiments, $\epsilon = 0.025$ and histograms have $256$ bins).
To produce our final histogram features $N$ take the square root of the L1-normalized histogram counts, which generally improves the effectiveness of histogram features \cite{Arandjelovic2012}.
\begin{equation}
N(u,v) = \sqrt{ M(u,v) \over \sum_{u',v'} M(u',v') }
\end{equation}
Any normalization or transformation is allowed at this step as long as the same operation is applied to the entire histogram, though at other stages in the algorithm care must be taken to preserve translational invariance.

\begin{figure}[!]
\centering
  \begin{subfigure}[!]{1.05in}
    \includegraphics[width=1.05in]{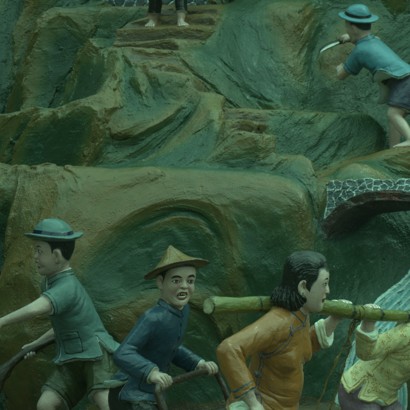} \\
    \includegraphics[width=1.05in]{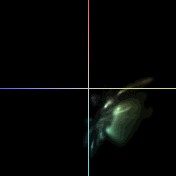}
    \caption{Input Image \label{fig:tints1}}
  \end{subfigure}
  \begin{subfigure}[!]{1.05in}
    \includegraphics[width=1.05in]{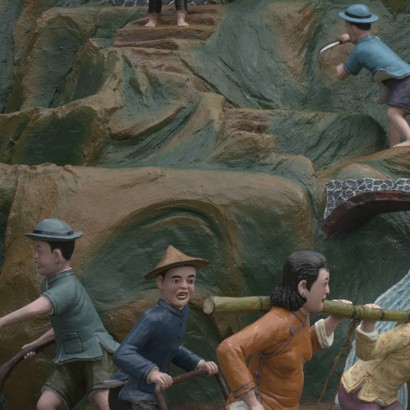} \\
    \includegraphics[width=1.05in]{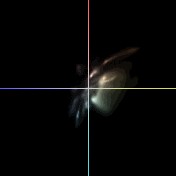}
    \caption{True Image \label{fig:tints2}}
  \end{subfigure}
  \begin{subfigure}[!]{1.05in}
    \includegraphics[width=1.05in]{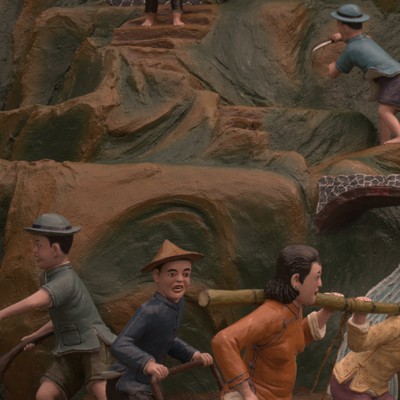} \\
    \includegraphics[width=1.05in]{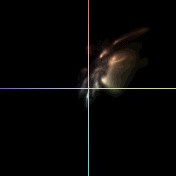}
    \caption{Tinted Image \label{fig:tints3}}
  \end{subfigure}
  \caption{Some images and their log-chrominance histograms (with an axis overlayed for easier visualization, horizontal = $u$, vertical = $v$).
  The images are the same except for ``tints'' --- scaling of red and blue.
  Tinting an image affects the image's histogram only by a translation in log-chrominance space.
  This observation enables our convolutional approach to color correction, in which our algorithm learns to localize a histogram in this 2D space.
  \label{fig:tints}
  }
\end{figure}

In Figure~\ref{fig:tints} we show three tinted versions of the same image with each image's chrominance histogram.
Note that each histogram is a translated version of the other histograms (ignoring sampling artifacts) and that the shape of the histogram does not change.
This is a consequence of our definitions of $u$ and $v$: scaling a pixel's RGB value is equivalent to translating a pixel's log-chrominance, as was noted in \cite{FinlaysonJOSA2001}.
This equivalence between image-tinting and histogram-shifting enables the rest of our algorithm.

Our algorithm works by considering all possible tints of an image, scoring each tinted image, and then returning the highest-scoring tint as the estimated illumination of the input image.
This may sound like an expensive proposition as it requires a brute-force search over all possible tints, where some scoring function is applied at each tint.
However, provided that the scoring function is a linear combination of histogram bins, this brute-force search is actually just the convolution of $N$ with some filter $F$, and there are many ways that convolution can be made efficient.
This gives us a sketch of our algorithm: we will construct a histogram $N$ from the input image $I$, convolve that histogram with some filter $F$, and then use the highest-scoring illumination $\hat L$ to produce $\hat W = I / \hat L$.
More formally:
\begin{equation}
(\hat L_u, \hat L_v) = \underset{u,v}{\arg\max} \left( N * F \right)
\end{equation}
A visualization of this procedure (actually, a slightly more complicated version which will be explained later) can be seen in Figure~\ref{fig:pipeline}.
Now we require a way to learn a filter $F$ from training data such that this convolution produces accurate output.

\begin{figure}[!]
\centering
  \includegraphics[width=1.05in]{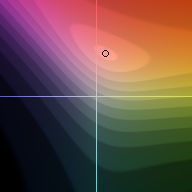}
  \includegraphics[width=1.05in]{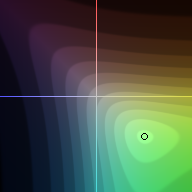}
  \includegraphics[width=1.05in]{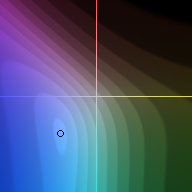}
  \caption{Visualizations of the cost function used during training $C\left(u, v, u^*, v^*\right)$ as a function of the proposed illumination color $(u,v)$, with each plot showing a different choice of the ground-truth illumination color $(u^*, v^*)$ (circled).
  Darker luminance means higher cost.
  These cost functions are used during training to encourage our learned filter to ``fire'' strongly at the true illuminant $(u^*, v^*)$ when convolved with the input histogram.
  \label{fig:costs}
  }
\end{figure}

To learn $F$ we use a model similar to multinomial logistic regression or structured prediction, in a convolutional framework.
Formally, our optimization problem is:
\begin{align}
& \min_F \, \lambda \sum_{u,v} F(u,v)^2 + \sum_{i,u,v} P(u,v) \,C\!\left(u,v,L^{(i)}_u, L^{(i)}_v \right) \nonumber \\
& P(u,v) = { \exp\left( (N^{(i)} * F)(u,v) \right) \over \sum_{u',v'} \exp\left( (N^{(i)} * F)(u',v') \right) }
\label{eq:loss}
\end{align}
Where $F$ is the filter whose weights we learn, $\{ N^{(i)} \}$  and $\{ L^{(i)} \}$ are our training-set chrominance histograms and ground-truth illuminations, respectively, and $(N^{(i)} * F)(u,v)$ is the convolution of $N^{(i)}$ and $F$ indexed at location $(u,v)$.
For convenience we define $P(u,v)$ which is a softmax probability for each $(u,v)$ bin in our histogram as a function of $N^{(i)} * F$.
We regularize our filter weights by minimizing the sum of squares of the elements of $F$, moderated by some hyperparameter $\lambda$.
At a high level, minimizing our loss finds an $F$ such that $N^{(i)} \ast F$ is larger at $(L^{(i)}_u, L^{(i)}_v)$ than it is elsewhere, where $C(u,v,u^*,v^*)$ defines the loss incurred at mis-estimated illuminants:
\begin{align}
C\left(u, v, u^*, v^*\right) &= \arccos \left({  \langle \ell,  \ell^* \rangle \over \norm{\ell} \, \norm{\ell^*} } \right) \nonumber \\
\ell =& [\exp(-u), \,\, 1, \,\, \exp(-v)]^\mathrm{T} \nonumber \\
\ell^* =& [\exp(-u^*), 1, \exp(-v^*)]^\mathrm{T}
\end{align}
$C$ measures the angle between the illuminations defined by $(u,v)$ and $(u^*, v^*)$, which is the error by which color-constancy algorithms are commonly evaluated.
Visualizations of $C$ can be seen in Figure~\ref{fig:costs}.
During training we initialize $F$ to all zeros (initialization does not appear to affect accuracy) and we minimize Eq.~\ref{eq:loss} first using a variant of stochastic gradient descent (detailed in supplement) followed by batch L-BFGS until convergence.
Using both optimization techniques produces lower losses and test-set error rates than using only SGD, but more quickly than only using batch L-BFGS.
Though our loss function is non-convex, optimization appears to work well and our learned model performs better than other models trained with various convex approximations to our loss function.

Our problem resembles multinomial logistic regression, but where every $(u,v)$ has a variable loss $C$ measuring the cost of each possible $(u,v)$ chrominance with respect to some ground-truth chrominance $(u^*, v^*)$.
The use of a softmax makes our model resemble a classification problem, and the use of a variable cost makes our model resemble structured prediction.
We experimented with simply minimizing the cross-entropy of $P(u,v)$ with respect to a delta function at $(u^*, v^*)$, and with using maximum-margin structured prediction \cite{taskar2005} with margin rescaling and slack rescaling, but found that our proposed approach produced more accurate results on the test set.
We also experimented with learning a ``deep'' set of filters instead of a single filter $F$, thereby resulting in a convolutional neural network \cite{lecun1989backpropagation}, but we found the amount of training data in our datasets insufficient to prevent overfitting.

A core property of our approach is that our model is trained \emph{discriminatively}.
Our structured-prediction approach means that $F$ is learned directly in accordance with the criteria we care about --- how accurately it identifies each illumination color in the training set.
This is very different from the majority of color constancy algorithms which either learn or analytically construct generative models of the distributions of colors in natural images viewed under white light.
To demonstrate the importance of discriminative training, we will evaluate against a generatively-trained version of our model which learns a model to maximize the likelihood of colors in natural images, while not considering that this generative model will be used for a discriminative task.
Our generative model learns our filter $F$ according to the following optimization problem:
\begin{align}
& \max_F \, \sum_i \sum_{u,v} \left( \log\left(P(u,v) \right) N^{(i)}(u,v) \right)  \nonumber \\
& P(u,v) = { \exp\left( (\delta^{(i)} * F)(u,v) \right) \over \sum_{u',v'} \exp\left( (\delta^{(i)} * F)(u',v') \right) } \nonumber \\
& \delta^{(i)} = \left[ \left( \left|u - L^{(i)}_u \right| \leq \sfrac{\epsilon}{2} \right) \wedge \left( \left| v - L^{(i)}_v \right| \leq \sfrac{\epsilon}{2} \right) \right]
\label{eq:gen_loss}
\end{align}
Minimizing this loss produces a filter $F$ such that, when $F$ is convolved with a delta function located at the illuminant color's chrominance, the categorical distribution produced by exponentiating that filter output maximizes the likelihood of the training set chroma histograms $\{ N^{(i)} \}$.
We do not regularize $F$, as it does not improve performance when generative training is used.

\begin{figure}[!]
\centering
  \stackunder[5pt]
    {\includegraphics[width=1.6in]{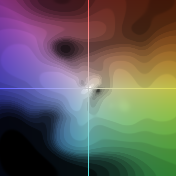}}
    {discriminative $F$}
    \stackunder[5pt]
    {\includegraphics[width=1.6in]{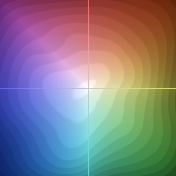}}
    {generative $F$}
  \caption{
  Learned filters on the same training data, with the left filter learned discriminatively and the right filter learned generatively.
  The generative model just learns a simple ``gray-world'' like filter, while the discriminative model learns to do things like upweight blues that resemble the sky and downweight pale greens that resemble badly white-balanced images.
  One can think of the discriminatively learned filter as a histogram of colors in well white-balanced images \emph{minus} a histogram of colors in poorly white-balanced images
  \label{fig:disc_gen}
  }
\end{figure}

A visualization of filters learned discriminatively and generatively on the same data can be seen in Figure~\ref{fig:disc_gen}.
We see that discriminative training learns a much richer and more elaborate model than the generative model.
This is because our discriminative training does not just learn what white-balanced images look like --- it learns how to distinguish between white-balanced images and improperly white-balanced images.
In Section~\ref{sec:results} we will show that discriminative training substantially improves model accuracy.

\section{Efficient Filtering}
\label{sec:filtering}

Though our algorithm revolves around a linear filter $F$ with which we will convolve our chroma histograms, the specific parametrization of $F$ affects the accuracy and speed of our model.
For example, a filter the size of the input histogram would be likely to overfit and would be expensive to evaluate.
We found that accurate filters for our task tend to have a log-polar or ``retinotopic'' structure, in which the filter contains a large amount of high-frequency variation near the center of the filter but only contains low-frequency variation far from the center.
Intuitively, this makes sense: when localizing the illumination color of an image, the model should pay close attention to chroma variation near the predicted white point, while only broadly considering chroma variation far from the predicted white point.

\begin{figure}[!]
\centering
  \includegraphics[width=3.2in]{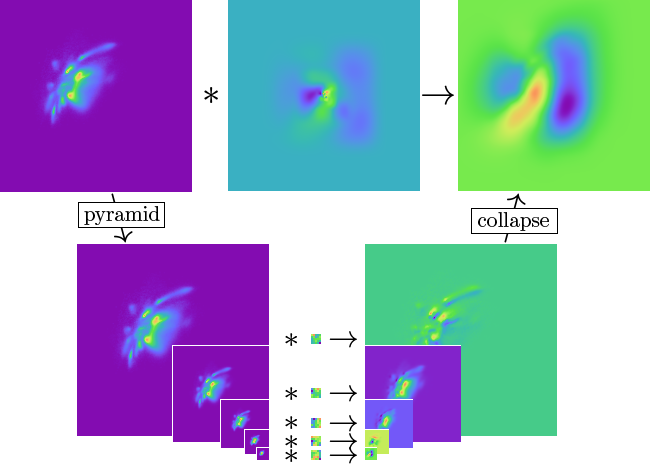}
  \caption{
  Here we visualize the ``pyramid filter'' \cite{BarronICCV2013} used to score chroma histograms.
  Above we show naive convolution of a histogram (top left) with a retina-like filter (top middle), while below we evaluate that same filter more efficiently by constructing a pyramid from the histogram, convolving each scale of the pyramid with a small filter, and then collapsing the filtered histogram.
  By using the latter filtering approach we simplify regularization during training and improve speed during testing.
  \label{fig:pyrconv}
  }
\end{figure}

With the goal of a fast retina-like filter, we chose to use the ``pyramid filtering'' technique of \cite{BarronICCV2013} for our histogram convolution.
Pyramid filtering works by first constructing a Gaussian pyramid of the input signal (in this case, we construct a $7$-level pyramid from $N(u,v)$ using bilinear downsampling), then filtering each scale with a small filter (we used $5 \times 5$ filters), and then collapsing the filtered pyramid down into an image (using bilinear upsampling).
When collapsing the pyramid we found it necessary to apply a $[1, 2, 1]$ blur before each upsample operation to reduce sampling artifacts.
This filter has several desirable properties: it is efficient to compute, there are few free parameters so optimization and regularization are easy, and the filter can describe fine detail in the center while modeling coarse context far from the center.
We regularize this filter by simply minimizing the squared 2-norm of the filter coefficients at each scale, all modulated by a single hyperparameter $\lambda$, as in Eq.~\ref{eq:loss} (this is actually a slight departure from Eq.~\ref{eq:loss} as the regularization is now in a linearly transformed space).
A visualization of pyramid filtering can be seen in Figure~\ref{fig:pyrconv}.

As described in \cite{BarronICCV2013} pyramid filtering is equivalent to, for every pixel, computing a feature with a log-polar sampling pattern and then classifying that feature with a linear classifier.
This sort of feature resembles standard features used in computer vision, like shape context \cite{Belongie00}, geometric blur \cite{Berg2001}, FREAK features \cite{Freak2012}, DAISY \cite{Tola10}, etc.
However, the pyramid approximation requires that the sampling pattern of the feature be rectangular instead of polar, that the scales of the feature be discretized to powers of 2, and that the sampling patterns of the feature at each scale overlap.
This difference makes it tractable to compute and classify these features densely at every pixel in the image, which in turn allows us to estimate the illuminant color very precisely.

\section{Generalization}
\label{sec:channels}

The previously described algorithm can estimate the illumination $L$ from an image $I$ by filtering a histogram $N$ constructed from the chroma values of the pixels in $I$.
Effectively, this model is a sophisticated kind of ``gray world'' algorithm, in that all spatial information is ignored and the image is treated like a ``bag'' of pixels.
However, well-performing color constancy algorithms generally use additional sources of information, such as the color of edges \cite{Barnard00,Finlayson2013,vandeWeijerTIP2007} or spatial neighborhoods \cite{Chakrabarti11}.
To that end, we present an extension of our algorithm in which instead of constructing and classifying a single histogram $N$ from a single image $I$, we filter a set of histograms $\{ N_j \}$ from a set of ``augmented'' images $\{ I'_j \}$, and sum the filtered responses before computing softmax probabilities.
These augmented images will reflect edge and spatial statistics of the image $I$, thereby enabling our model reason about multiple sources of chroma information beyond individual pixel chroma.

Naively one might attempt to construct these augmented images $\{ I'_j \}$ by simply applying common image processing operations to $I$, such as applying a filter bank, median filters, morphological operations, etc.
But remember from Section~\ref{sec:problem} that the image from which we construct chroma histograms must exactly map scaling to the channels of the input image to shifts in chroma histogram space.
This means that our augmented images must also map a per-channel scaling to the same shift in histogram space, limiting the set of possible augmented images that we can use.

\begin{figure}[!]
\centering
    \stackunder[5pt]{
      \includegraphics[width=0.8in]{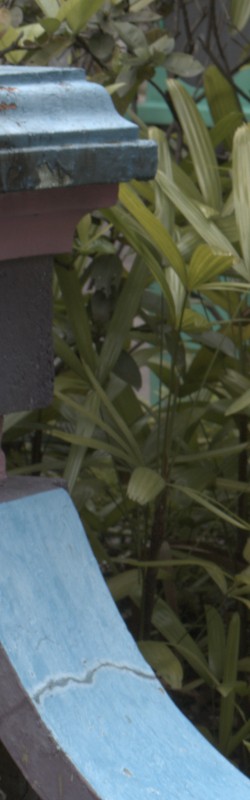}
      }{$I'_1 = I$}
      \hspace{-0.15in}
    \stackunder[5pt]{
      \includegraphics[width=0.8in]{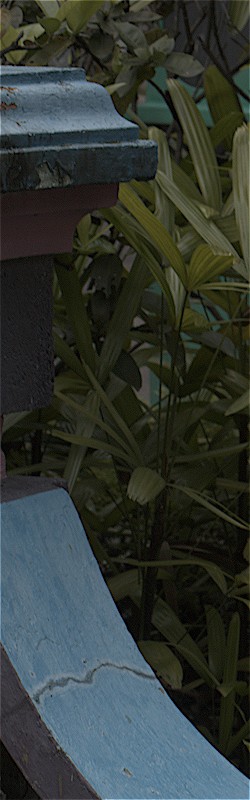}
      }{$I'_2$}
      \hspace{-0.15in}
    \stackunder[5pt]{
      \includegraphics[width=0.8in]{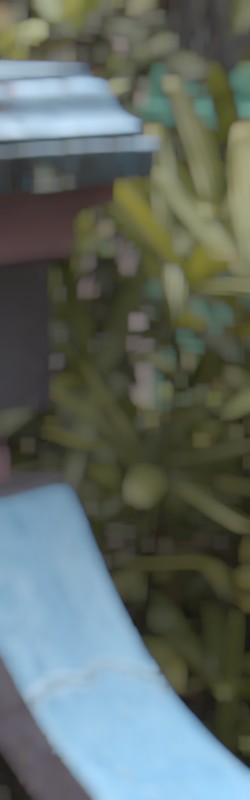}
      }{$I'_3$}
      \hspace{-0.15in}
    \stackunder[5pt]{
      \includegraphics[width=0.8in]{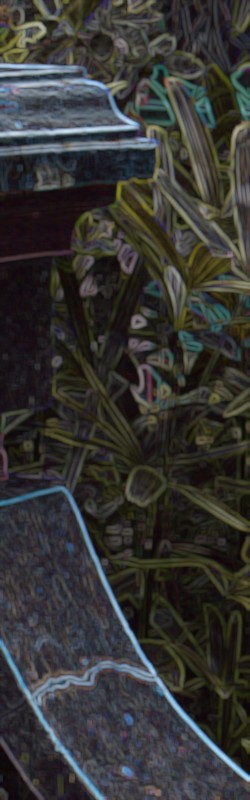}
      }{$I'_4$}
    \caption{
    Though our model can take the pixel values of the input image $I$ as its sole input, performance can be improved by using a set of ``augmented'' images $\{ I' \}$.
    Our extended model uses three augmented images which capture local spatial information (texture, highlights, and edges, respectively) in addition to the input image.
    \label{fig:channels}
}
\end{figure}

For our color-scaling/histogram-shifting requirement to be met, our augmented-image mappings must preserve scalar multiplication: a scaled-then-filtered version of a channel in the input image $I$ must be equal to a filtered-then-scaled version of that channel.
This problem is alluded to in \cite{Finlayson2013}, in which the authors limit themselves to ``color moments which scale with intensity''.
Additionally, the output of the mappings must be non-negative as we will need to compute the logarithm of the output of each mapping (the input is assumed to be non-negative).
Here are three mappings which satisfy our criteria:
\begin{align}
f(I, \mathit{filt} ) &= \max(0, I * \mathit{filt}) \nonumber \\
g(I, \rho, w) &= \mathrm{blur}( I^\rho, w )^{1 / \rho} \nonumber \\
h(I, \rho, w) &= \left( \mathrm{blur}( I^\rho, w ) - \mathrm{blur}( I , w)^\rho \right)^{1 / \rho}
\end{align}
Where $\mathrm{blur}( \cdot, w )$ is a box filter of width $w$.
$f(\cdot, \mathit{filt})$ convolves each channel of the image with some filter $\mathit{filt}$ and then clamps the filtered value to be at least $0$.
$g(\cdot, p, w)$ computes a local norm of pixel values in $I$ such that $g(\cdot, 1, w)$ is a blur, $g(\cdot, \infty, w)$ is a ``max'' filter, and $g(\cdot, -\infty, w)$ is a ``min'' filter.
$h(\cdot)$ computes a kind of normalized moment of pixel values, where $h(\cdot, 2, w)$ is the local standard deviation of pixel values --- an unoriented edge/texture detector.
These operations all preserve scalar multiplication:
\begin{align}
f(\alpha I, \mathit{filt}) &= \alpha  f(I, \mathit{filt}) \nonumber \\
g(\alpha I, \rho, w) &= \alpha  g(I, \rho, w) \nonumber \\
h(\alpha I, \rho, w) &= \alpha  h(I, \rho, w)
\end{align}
In our extended model we use four augmented images: the input image $I$ itself, a ``sharpened'' and rectified $I$, a ``soft'' max-filtered $I$, and a standard-deviation-filtered $I$.
\begin{align}
I'_1 &= I \nonumber \\
I'_2 &= \max\left(0, I * \tiny \left[  \begin{array}{@{}r@{\,\,}r@{\,\,}r@{\,}}
0 & -1 & 0 \\
-1 & 5 & -1 \\
0 & -1 & 0 \end{array} \right] \right) \nonumber \\
I'_3 &= \mathrm{blur}( I^4, 11 )^{1 / 4} \nonumber \\
I'_4 &= \sqrt{ \mathrm{blur}( I^2, 3) - \mathrm{blur}( I , 3)^2 }
\end{align}
Other similar channels or compositions of these channels could be used as well, though we use a small number of simple channels here for the sake of speed and to prevent overfitting.
See Figure~\ref{fig:channels} for visualizations of the information captured by each of these channels.
During training we simply learn $4$ pyramid filters instead of $1$ and sum the individual filter responses before computing the softmax probabilities in Eq.~\ref{eq:loss}.

\begin{figure*}[!]
\centering
  \stackunder[5pt]
    {\includegraphics[width=1.7in]{figures/pipeline_I.jpg}}
    {$I$}
  \stackunder[5pt]
    {\includegraphics[width=0.42in]{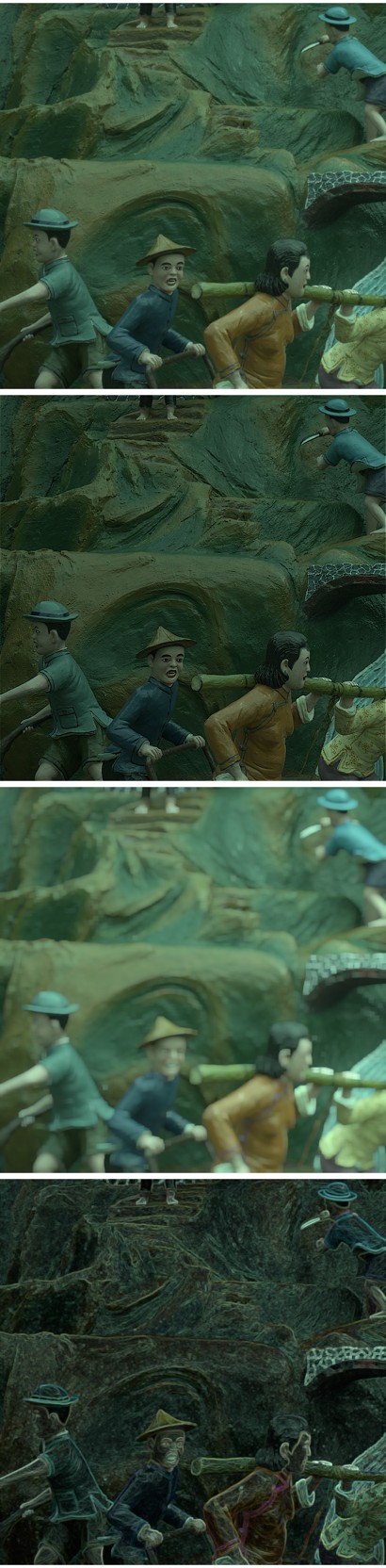}}
    {$\{ I'_j \} $}
  \stackunder[5pt]
    {\includegraphics[width=0.42in]{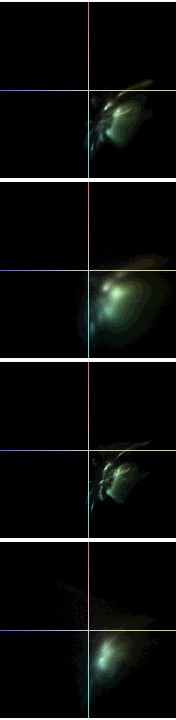}}
    {$\{ N_j \} $}
  \stackunder[5pt]
    {\includegraphics[width=0.42in]{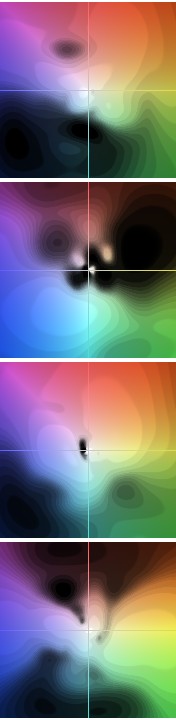}}
    {$\{ F_j \} $}
  \stackunder[5pt]
    {  \includegraphics[width=1.7in]{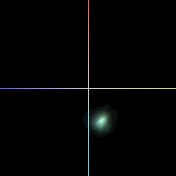}}
    {$\exp(\sum_j F_j \ast N_j) $}
  \stackunder[5pt]
    {\includegraphics[width=1.7in]{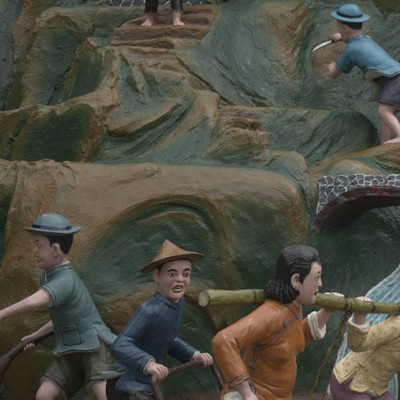}
    \hspace{-0.07in}
     \includegraphics[width=0.212in]{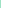}}
    { \quad output: \quad $\hat W$ \,\,\qquad\qquad\,\,\,\, $\hat L$}
  \caption{
  An overview of inference in our model for a single image.
  An input image $I$ is transformed into a set of scale-preserving augmented images $\{ I'_j \}$ which highlight different aspects of the image (edges, patches, etc).
  The set of augmented images is turned into a set of chroma histograms $\{ N_j \}$, for which we have learned a set of weights in the form of pyramid filters $\{ F_j \}$.
  The histograms are convolved with the filters and then summed, giving us a score for all bins in our chroma histogram.
  The highest-scoring bin is assumed to be the color of the illuminant $\hat L$, and the output image $\hat W$ is produced by dividing the input image by that illuminant.
  \label{fig:pipeline}
  }
\end{figure*}

Now that all of our model components have been defined, we can visualize inference in our final model in Figure~\ref{fig:pipeline}.

\section{Results}
\label{sec:results}

We evaluate our algorithm on two datasets: the Color Checker Dataset \cite{Gehler08} reprocessed by Shi and Funt \cite{shifunt}, and the dataset from Cheng \etal~\cite{Cheng14}.
The Color Checker dataset is widely used and is reasonably large --- $568$ images from a single camera.
The dataset from Cheng \etal~is larger, with $1736$ images taken from $8$ different cameras, but the same scene is imaged multiple times by each of the 8 cameras.
As is standard, we evaluate using three-fold cross-validation, computing the angle in degrees between our estimated illumination $\hat L$ and the true illumination $L^*$ for each image.
We report several statistics of these errors: the mean, the median, the tri-mean, the means of the errors in the lowest-error $25\%$ of the data and the highest-error $25\%$ of the data, and for the Color Checker Dataset the $95$th percentile.
Some baseline results on the Color Checker Dataset were taken from past papers, thereby resulting in some missing error metrics for some algorithms.
For the Cheng \etal~dataset we also report an average error, which is the geometric mean of the other error statistics.

Cheng \etal~ran $8$ different experiments with their $8$ different cameras, which makes tersely summarizing performance difficult.
To that end, we report the geometric mean of each error metric for each algorithm across all cameras.
We computed results for our own algorithm identically: we learn a model for each camera independently, compute errors for each camera, and then report the geometric mean across all cameras.

Our results can be seen in Tables~\ref{table:color_checker} and \ref{table:cheng}.
On the Color Checker Dataset we see a $30\%$ and $39\%$ reduction in error (mean and median, respectively) from the state-of-the-art (``Corrected-Moment''~\cite{Finlayson2013}), and on the dataset of Cheng~\etal~we see a $22\%$ reduction in average error from the state-of-the-art (Cheng~\etal).
This improvement is fairly consistent across different choices of error metrics.
The increased improvement on the Color Checker Dataset is likely due to the larger size of the Color Checker Dataset ($\sim\!379$ training images as opposed to $\sim\!144$, for three-fold cross validation), which likely favors our learning-based approach.
An example of our performance with respect to the state of the art can be seen in Figure~\ref{fig:problem} and in the supplement.

In our experiments we evaluated several different versions of our algorithm (``CCC''), indicated by the name of each model.
Models labeled ``gen'' are trained in a generative fashion (Eq.~\ref{eq:gen_loss}), while ``disc'' models are trained discriminatively (Eq.~\ref{eq:loss}).
Models labeled ``simp'' use our simple feature set (just the input image) while ``ext'' models use the four augmented images from Section~\ref{sec:channels}.
Our results show that discriminative training is superior to generative training by a large margin ($30-40\%$ improvement), and that using our extended model produces better results than our simple model ($10-20\%$ improvement).
Our generatively-trained models perform similarly to some past techniques which were also trained in a generative fashion, suggesting that the use of discriminative training is the driving force behind our algorithm's performance.

Though most of our baseline results were taken from past papers, to ensure a thorough and fair evaluation we obtained the code for the best-performing technique on the Color Checker dataset (``Corrected-Moment''~\cite{Finlayson2013}) and ran it ourselves on the Cheng~\etal\ dataset.
We also ran this code on the Color Checker dataset and reported our reproduced results, which differ slightly from those reported in \cite{Finlayson2013} apparently due to different parameter settings or inconsistencies between the provided code and the paper \cite{FinlaysonComm}.
Results for the corrected moment algorithm produced by ourselves are indicated with asterisks in Tables~\ref{table:color_checker} and \ref{table:cheng}.

Evaluating our trained model is reasonably fast.
With our unoptimized Matlab implementation running on a 2012 HP Z420 workstation, for each image it takes about $1.2$ seconds per megapixel to construct our augmented images and produce normalized log-chrominance histograms from them, and about 20 milliseconds to pyramid-filter those histograms and extract the argmax.

\section{Conclusion}
\label{sec:conclusion}

We have presented CCC, a novel learning-based algorithm for color constancy.
Our technique builds on the observation that the per-channel scaling in an image caused by the color of the illumination produces a translation in the space of log-chroma histograms.
This observation lets us leverage ideas from object detection and structured prediction to discriminatively train a convolutional classifier to perform well at white balancing, as opposed to the majority of prior work which uses generative training.
Our algorithm is made more efficient by using a pyramid-based approach to image filtering, and is made more accurate by augmenting the input to our algorithm with variants of the input image that capture different kinds of spatial information.
Our technique produces state-of-the-art performance on the two largest color constancy datasets, beating the best-performing techniques by $20\%-40\%$ on various error metrics.
Our experiments suggest that color constancy algorithms may benefit from much larger datasets than are currently used, as has been the case for object detection and recognition.
This newly-established connection with object detection suggests that color constancy may be a fruitful domain for researchers to apply new object detection techniques.
Furthermore, our results show that many of the lessons learned from discriminative machine learning are more relevant to color constancy than has been previously thought, and suggests that other core low-level vision and imaging tasks may benefit from a similar reevaluation.

% FUCK add the CNN paper

\begin{table}[!]
\begin{center}
\resizebox{3.25in}{!}{
\Huge
\begin{tabular}{ l  | c  c  c c c c}
\multirow{2}{*}{Algorithm} & \multirow{2}{*}{Mean} & \multirow{2}{*}{Med.} & \multirow{2}{*}{Tri.} & Best & Worst & 95\% \\
 & & & & 25\% & 25\% & Quant. \\
\hline
White-Patch \cite{Brainard86} & $7.55$ & $5.68$ & $6.35$ & $1.45$ & $16.12$ & - \\
Edge-based Gamut \cite{Barnard00} & $6.52$ &  $5.04$ & $5.43$ & $1.90$ & $13.58$ & - \\
Gray-World \cite{Buchsbaum80} & $6.36$ & $6.28$ & $6.28$ & $2.33$ & $10.58$ & $11.3$ \\
1st-order Gray-Edge \cite{vandeWeijerTIP2007} & $5.33$ & $4.52$ & $4.73$ & $1.86$ & $10.03$ & $11.0$  \\
2nd-order Gray-Edge \cite{vandeWeijerTIP2007} & $5.13$ & $4.44$ & $4.62$ & $2.11$ & $9.26$ & - \\
Shades-of-Gray \cite{FinlaysonT04} & $4.93$ & $4.01$ & $4.23$ & $1.14$ & $10.20$ & $11.9$ \\
Bayesian \cite{Gehler08} & $4.82$ & $3.46$ & $3.88$ & $1.26$ & $10.49$ & - \\
General Gray-World \cite{Barnard02} & $4.66$ & $3.48$ & $3.81$ & $1.00$ & $10.09$ & - \\
Intersection-based Gamut \cite{Barnard00} & $4.20$ & $2.39$ & $2.93$ & $0.51$ & $10.70$ & - \\
Pixel-based Gamut \cite{Barnard00} & $4.20$ & $2.33$ & $2.91$ & $0.50$ & $10.72$ & $14.1$ \\
Natural Image Statistics \cite{GijsenijTPAMI2011} & $4.19$ & $3.13 $ & $ 3.45 $ & $ 1.00 $ & $ 9.22$ & $11.7$  \\
Bright Pixels \cite{Hamid12} & $3.98$ & $2.61$ & - & - & - & - \\
Spatio-spectral (GenPrior) \cite{Chakrabarti11} & $3.59 $ & $ 2.96 $ & $ 3.10 $ & $ 0.95 $ & $ 7.61$ & - \\
Cheng \etal~\cite{Cheng14} & $3.52$ & $2.14$ &  $2.47$ &  \cellcolor{Yellow} $0.50$ & $8.74$ & - \\
Corrected-Moment (19 Color) \cite{Finlayson2013} & $3.5$ & $2.6$ & - & - & - & $8.6$ \\
Corrected-Moment (19 Edge) \cite{Finlayson2013} & \cellcolor{Yellow} $2.8$ & \cellcolor{Yellow} $2.0$ & - & - & - & \cellcolor{Yellow}  $6.9$ \\
Corrected-Moment* (19 Color) \cite{Finlayson2013} & $2.96$ & $2.15$ & \cellcolor{Yellow} $2.37$ & $0.64$ & $ 6.69$ & $8.23$ \\
Corrected-Moment* (19 Edge) \cite{Finlayson2013} & $3.12$ & $2.38$ & $2.59$ & $0.90$ & \cellcolor{Yellow}  $6.46$ & $7.80$ \\
% Bianco \etal \cite{Bianco2015} & $ 2.63 $ & $1.98$ & - & - & - \\
\hline
     CCC (gen+simp) & $ 3.57 $ & $ 2.62 $ & $ 2.90 $ & $ 1.01 $ & $ 7.74 $ & $ 9.62 $\\
      CCC (gen+ext) & $ 3.24 $ & $ 2.33 $ & $ 2.61 $ & $ 1.02 $ & $ 6.88 $ & $ 8.42 $\\
    CCC (disc+simp) & $ 2.48 $ & $ 1.52 $ & $ 1.70 $ & $ 0.37 $ & $ 6.26 $ & $ 8.30 $\\
     CCC (disc+ext) & \cellcolor{Red}$ 1.95 $ & \cellcolor{Red}$ 1.22 $ & \cellcolor{Red}$ 1.38 $ & \cellcolor{Red}$ 0.35 $ & \cellcolor{Red}$ 4.76 $ & \cellcolor{Red}$ 5.85 $
\end{tabular}
}
\vspace{1mm}
\caption{
Performance on the reprocessed \cite{shifunt} Color Checker Dataset \cite{Gehler08}.
For each metric the best-performing technique is highlighted in red, and the second-best-performing (excluding variants of our technique) is in yellow.
Some baseline numbers here were taken from past work  \cite{Cheng14, GijsenijTIP2011}, which used different error measures, thereby resulting in some missing entries.
\label{table:color_checker}}
\end{center}
\end{table}

\begin{table}[!]
\begin{center}
\resizebox{3.25in}{!}{
\Huge
\begin{tabular}{ l  | c c c c c | c}
\multirow{2}{*}{Algorithm} & \multirow{2}{*}{Mean} & \multirow{2}{*}{Med.} & \multirow{2}{*}{Tri.} & Best & Worst & \multirow{2}{*}{Avg.} \\
 & & & & 25\% & 25\% &  \\
\hline
                    White-Patch \cite{Brainard86}      & $ 10.62 $ & $ 10.58 $ & $ 10.49 $ & $ 1.86 $ & $ 19.45 $ & $ 8.43 $ \\
                Edge-based Gamut \cite{Barnard00}      & $ 8.43 $ & $ 7.05 $ & $ 7.37 $ & $ 2.41 $ & $ 16.08 $ & $ 7.01 $ \\
               Pixel-based Gamut \cite{Barnard00}      & $ 7.70 $ & $ 6.71 $ & $ 6.90 $ & $ 2.51 $ & $ 14.05 $ & $ 6.60 $ \\
        Intersection-based Gamut \cite{Barnard00}      & $ 7.20 $ & $ 5.96 $ & $ 6.28 $ & $ 2.20 $ & $ 13.61 $ & $ 6.05 $ \\
                    Gray-World \cite{Buchsbaum80}      & $ 4.14 $ & $ 3.20 $ & $ 3.39 $ & $ 0.90 $ & $ 9.00 $ & $ 3.25 $ \\
                         Bayesian \cite{Gehler08}      & $ 3.67 $ & $ 2.73 $ & $ 2.91 $ & $ 0.82 $ & $ 8.21 $ & $ 2.88 $ \\
Natural Image Statistics \cite{GijsenijTPAMI2011}      & $ 3.71 $ & $ 2.60 $ & $ 2.84 $ & $ 0.79 $ & $ 8.47 $ & $ 2.83 $ \\
               Shades-of-Gray \cite{FinlaysonT04}      & $ 3.40 $ & $ 2.57 $ & $ 2.73 $ & $ 0.77 $ & $ 7.41 $ & $ 2.67 $ \\
        Spatio-spectral (ML) \cite{Chakrabarti11}      & $ 3.11 $ & $ 2.49 $ & $ 2.60 $ & $ 0.82 $ & $ 6.59 $ & $ 2.55 $ \\
              General Gray-World \cite{Barnard02}      & $ 3.21 $ & $ 2.38 $ & $ 2.53 $ & $ 0.71 $ & $ 7.10 $ & $ 2.49 $ \\
    2nd-order Gray-Edge \cite{vandeWeijerTIP2007}      & $ 3.20 $ & $ 2.26 $ & $ 2.44 $ & $ 0.75 $ & $ 7.27 $ & $ 2.49 $ \\
                     Bright Pixels \cite{Hamid12}      & $ 3.17 $ & $ 2.41 $ & $ 2.55 $ & $ 0.69 $ & $ 7.02 $ & $ 2.48 $ \\
    1st-order Gray-Edge \cite{vandeWeijerTIP2007}      & $ 3.20 $ & $ 2.22 $ & $ 2.43 $ & $ 0.72 $ & $ 7.36 $ & $ 2.46 $ \\
  Spatio-spectral (GenPrior) \cite{Chakrabarti11}      & $ 2.96 $ & $ 2.33 $ & $ 2.47 $ & $ 0.80 $ & \cellcolor{Yellow} $ 6.18 $ & $ 2.43 $ \\
 Corrected-Moment* (19 Edge) \cite{Finlayson2013}      & $ 3.03 $ & $ 2.11 $ & $ 2.25 $ & $ 0.68 $ & $ 7.08 $ & $ 2.34 $ \\
Corrected-Moment* (19 Color) \cite{Finlayson2013}      & $ 3.05 $ & \cellcolor{Yellow} $ 1.90 $ & \cellcolor{Yellow} $ 2.13 $ & $ 0.65 $ & $ 7.41 $ & $ 2.26 $ \\
                       Cheng \etal \cite{Cheng14}      & \cellcolor{Yellow} $ 2.92 $ & $ 2.04 $ & $ 2.24 $ & \cellcolor{Yellow} $ 0.62 $ & $ 6.61 $ & \cellcolor{Yellow} $ 2.23 $ \\
\hline
  CCC (gen+simp) & $ 3.42 $ & $ 2.66 $ & $ 2.84 $ & $ 0.98 $ & $ 7.09 $ & $ 2.82 $ \\
  CCC (gen+ext) & $ 3.05 $ & $ 2.27 $ & $ 2.49 $ & $ 0.92 $ & $ 6.43 $ & $ 2.52 $ \\
CCC (disc+simp) & $ 2.61 $ & $ 1.70 $ & $ 1.87 $ & $ 0.52 $ & $ 6.28 $ & $ 1.94 $ \\
 CCC (disc+ext) & \cellcolor{Red}$ 2.38 $ & \cellcolor{Red}$ 1.48 $ & \cellcolor{Red}$ 1.69 $ & \cellcolor{Red}$ 0.45 $ & \cellcolor{Red}$ 5.85 $ & \cellcolor{Red}$ 1.74 $ \\
\end{tabular}
}
\vspace{1mm}
\caption{
Performance on the dataset from Cheng \etal~\cite{Cheng14}.
For each metric the best-performing technique is highlighted in red, and the second-best-performing (excluding variants of our technique) is in yellow.
\label{table:cheng}}
\end{center}
\end{table}

{\small
\bibliographystyle{ieee}
\bibliography{white_balance}
}

\end{document}

%%%%%%%%% TITLE

\title{Convolutional Color Constancy \\ Supplemental Material}

\author{Jonathan T. Barron\\
{\tt\small barron@google.com}
}

\maketitle

\section{Exponential Decay SGD}

Though we use conventional (batch) L-BFGS to finalize optimization when training our model, optimization can be sped up by using stochastic (non-batch) gradient descent techniques prior to L-BFGS. Inspired by recent advances in ``second-order'' stochastic gradient descent techniques \cite{Duchi2011, RMSProp, Zeiler2012}, we developed a novel variant of stochastic gradient descent based on exponential decaying different quantities at different rates, which we found to work well on our task.

Pseudocode for our ``exponential decay SGD'' technique can be seen in Algorithm~\ref{alg:sgd}. This technique is similar to RMSProp \cite{RMSProp} and AdaDelta \cite{Zeiler2012}, in that we maintain a moving average model of the gradient-squared, and then divide the gradient by the square-root of the average-gradient-squared before taking a gradient-descent step. But in addition to maintaining a moving average gradient-squared using exponential decay, we also maintain a moving average estimate of the gradient and of the loss. The moving average gradient serves as an alternative to commonly used ``mini-batches'', where instead of computing the average gradient of $n$ datapoints before taking a gradient descent step, we compute $n$ different gradients and take $n$ different gradient descent steps while averaging in past gradient estimates to prevent dramatic jumps during optimization. This appears to help optimization, especially in our domain where our training set sizes are somewhat small.

Many SGD techniques use an adaptive learning rate, where the learning rate is increased every epoch if optimization succeeds (ie, the loss decreases) and the learning rate is decreased if optimization fails (ie, the loss is increasing or oscillating). To generalize this idea we maintain a moving average of the loss for the entire dataset and compare every sampled datapoint's loss to that average. If a datapoint's loss appears is less than the average we slightly increase the step size, otherwise we slightly decrease the step size. In contrast to a per-epoch learning rate revision, this approach allows the step size to vary quickly during just a single epoch of optimization, thereby speeding up optimization.

We parametrized our technique in terms of half-life rather than using decay multipliers, which makes these parameters easier to reason about. For example, we found if effective to set the half-life for the average loss to be roughly the size of the dataset, so that the average loss is always a reflection of the entire dataset. The half-life of the gradient we set to be small --- about the size of a mini-batch, and the half-life of the gradient-squared we set to be significantly larger than that of the gradient but less than that of the loss.

To ensure that our running average estimates are correct even at the beginning of optimization, we model each moving average as the ratio of two quantities. Though we describe our algorithm as randomly sampling datapoints until convergence, in practice we optimize for a fixed number of epochs (in our experiments, $50$) and for each epoch we randomly order our data and then sample every datapoint in that random order, thereby improving the coverage of our training data.

\begin{algorithm*}[!]
\caption{Exponential Decay SGD \label{alg:sgd}}
{\bf Hyperparameters:} \\
\begin{tabular}{ll}
$\beta_{f} = 10^3$ & // The half-life of our exponentially decayed loss. \\
$\beta_{g} = 10^1$ & // The half-life of our exponentially decayed gradient. \\
$\beta_{h} = 10^2$ & // The half-life of our exponentially decayed gradient-squared. \\
$\alpha = 10^{-3}$  & // The initial step-size. \\
$\alpha^+ = 1.0001$ & // The amount to decrease the step-size when the loss decreases. \\
$\alpha^- = 0.999$ & // The amount to decrease the step-size when the loss increases. \\
$\epsilon = 10^{-5}$ & // A small constant to prevent divide-by-zero. \\
\end{tabular} \\
{\bf Input:} \\
\begin{tabular}{ll}
$\theta$ & // The initial model parameters. \\
$\{ X \}$ & // The training dataset. \\
$L(\cdot)$ & // The loss function
\end{tabular}
\\
\noindent\makebox[\linewidth]{\rule{\textwidth}{0.4pt}}
\begin{algorithmic}
\State $\lambda_f \gets \phantom{ 2^{-1/\lambda_f} }\mathllap{ 2^{-1/\lambda_f} }$ \quad // Convert the loss half-life to a decay.
\State $\lambda_g \gets \phantom{ 2^{-1/\lambda_g} }\mathllap{ 2^{-1/\lambda_g} }$ \quad // Convert the gradient half-life to a decay.
\State $\lambda_h \gets \phantom{ 2^{-1/\lambda_h} }\mathllap{ 2^{-1/\lambda_h} }$ \quad // Convert the gradient-squared half-life to a decay.
\State $f_n \gets 0, f_d \gets 0$ \quad // Initialize the moving average loss.
\State $g_n \gets 0, g_d \gets 0$ \quad // Initialize the moving average gradient.
\State $h_n \gets 0, h_d \gets 0$ \quad // Initialize the moving average gradient-squared.
\While{ not converged }
  \State $X_t \sim \{ X \}$ \quad // Sample a datapoint.
  \State $(f_t, g_t) \gets L\left(X_t, \theta \right)$ \quad // Compute the loss and gradient.

  \State $f_n \gets \lambda_f f_n + (1 - \lambda_f) f_t$ \quad // Decay the moving average numerators and add the new quantities.
  \State $g_n \gets \lambda_g g_n + (1 - \lambda_g) g_t$
  \State $h_n \gets \lambda_h h_n + (1 - \lambda_h) g_t^2$

  \State $f_d \gets \lambda_f f_d + (1 - \lambda_f) $  \quad // Decay the moving average  denominators and add the new mass.
  \State $g_d \gets \lambda_g g_d + (1 - \lambda_g)$
  \State $h_d \gets \lambda_h h_d + (1 - \lambda_f)$

  \State $\phantom{\bar f}\mathllap{ \bar f} = \phantom{h_n / h_d}\mathllap{f_n / f_d}$ \quad // Estimate the moving average loss.
  \State $\phantom{\bar f}\mathllap{ \bar g} = \phantom{h_n / h_d}\mathllap{g_n / g_d}$ \quad // Estimate the moving average gradient.
  \State $\phantom{\bar f}\mathllap{ \bar h} = \phantom{h_n / h_d}\mathllap{h_n / h_d}$ \quad // Estimate the moving average gradient-squared.
  \State $\displaystyle \Delta f = f_t - \bar f$ \quad // Compare the moving average loss to the current datapoint's loss.

  \If{ $\Delta f \leq 0$ } \quad // If the loss appears to be decreasing....
    \State{ $\alpha \gets \alpha^+ \times \alpha $} \quad // then increase the step size...
  \Else
    \State $\alpha \gets \alpha^- \times \alpha $ \quad // otherwise decrease the step size.
  \EndIf

  \State $\theta = \theta - \alpha \displaystyle { \bar g \over \sqrt{ \bar h + \epsilon^2}}$ \quad // Update the model parameters.
\EndWhile \\
\Return return $\theta$
\end{algorithmic}
\end{algorithm*}

\section{Additional Images}

To provide a better understanding of our results we present additional images from the reprocessed \cite{shifunt} Color Checker Dataset \cite{Gehler08}. For each image we show the input image produced by the camera and the ground-truth illumination and white-balanced image. All images are shown as cropped squares, for the sake of easy visualization. We also show the output of our model (``CCC''), and that of the three best-performing baseline techniques (two variants of the ``Corrected Moment'' algorithm \cite{Finlayson2013} and the technique of Cheng \etal \cite{Cheng14}. Though color checkers are visible in these images, the color checkers are cropped out of the images before being evaluated by any color constancy algorithm. To prevent ``cherry picking'', we sorted the $568$ images in the Color Checker Dataset by the average errors of the four algorithms being evaluated (ordering the images from ``easy'' to ``hard'') and evenly sampled images 1, 50, ... 551.

\begin{figure*}[b!]
\centering
  \stackunder[40pt]{
    \stackunder[10pt]{
    \stackunder[5pt]{Input image and ground-truth solution}
    {
    \stackunder[5pt]{
      \includegraphics[width=1.25in]{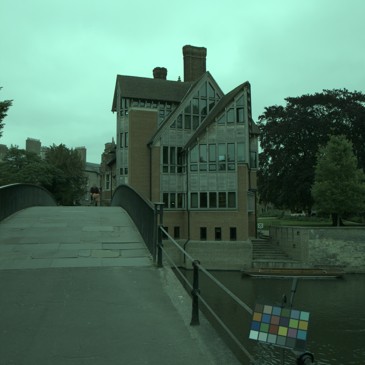}
      }{$I$}
    \stackunder[5pt]{
      \hspace{0.0in}
      }{$=$}
    \stackunder[5pt]{
      \includegraphics[width=1.25in]{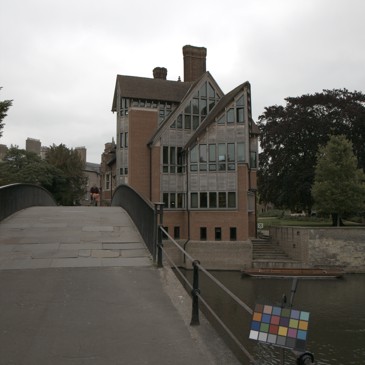}
    }{$W$}
    \stackunder[5pt]{
      \hspace{0.0in}
    }{$\times$}
    \stackunder[5pt]{
      \includegraphics[width=0.156in]{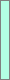}
      }{$L$}
    }}{
    \stackunder[5pt]{
     CCC
    }{
    \stackunder[5pt]{
    \includegraphics[width=1.25in]{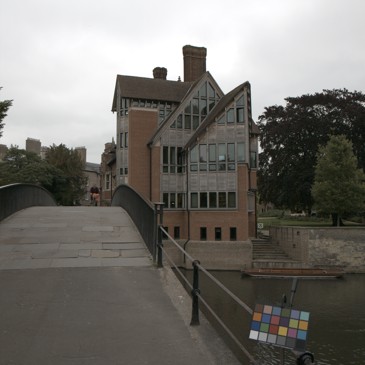}
    \hspace{-0.05in}
     \includegraphics[width=0.156in]{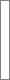}
    }
    {$\hat W$, $\hat L$, err = \input{figures/results/001_err_ours.txt}\textdegree}
    }
    \hspace{0.13in}
    \stackunder[5pt]{
     CM 19-Color \cite{Finlayson2013}
    }{
    \stackunder[5pt]{
    \includegraphics[width=1.25in]{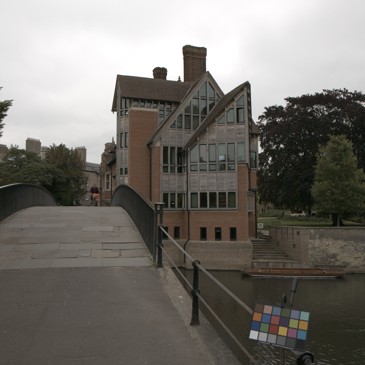}
    \hspace{-0.05in}
     \includegraphics[width=0.156in]{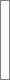}
     }
    { $\hat W$, $\hat L$, err = \input{figures/results/001_err_color.txt}\textdegree}
    }
    \hspace{0.13in}
    \stackunder[5pt]{
     CM 19-Edge \cite{Finlayson2013}
    }{
    \stackunder[5pt]{
    \includegraphics[width=1.25in]{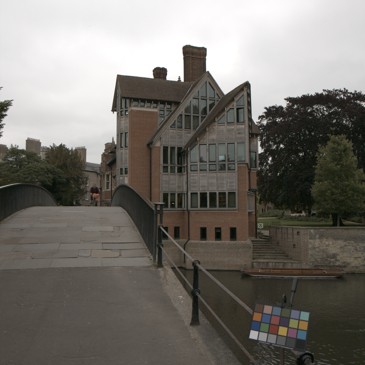}
    \hspace{-0.05in}
     \includegraphics[width=0.156in]{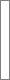}
     }
    { $\hat W$, $\hat L$, err = \input{figures/results/001_err_edge.txt}\textdegree}
    }
    \hspace{0.13in}
    \stackunder[5pt]{
     Cheng \etal, $p = 3.5$, \cite{Cheng14}
    }{
    \stackunder[5pt]{
    \includegraphics[width=1.25in]{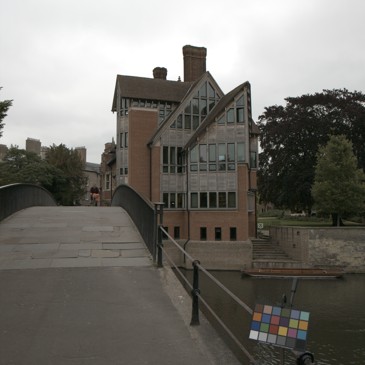}
    \hspace{-0.05in}
     \includegraphics[width=0.156in]{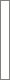}
     }
    { $\hat W$, $\hat L$, err = \input{figures/results/001_err_cheng.txt}\textdegree}
    }
    }
}{
    \stackunder[10pt]{
    \stackunder[5pt]{Input image and ground-truth solution}
    {
    \stackunder[5pt]{
      \includegraphics[width=1.25in]{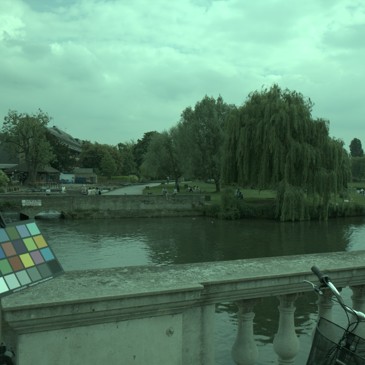}
      }{$I$}
    \stackunder[5pt]{
      \hspace{0.0in}
      }{$=$}
    \stackunder[5pt]{
      \includegraphics[width=1.25in]{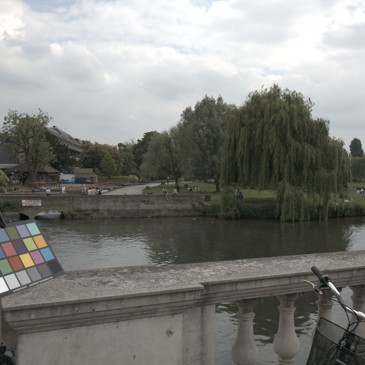}
    }{$W$}
    \stackunder[5pt]{
      \hspace{0.0in}
    }{$\times$}
    \stackunder[5pt]{
      \includegraphics[width=0.156in]{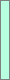}
      }{$L$}
    }}{
    \stackunder[5pt]{
     CCC
    }{
    \stackunder[5pt]{
    \includegraphics[width=1.25in]{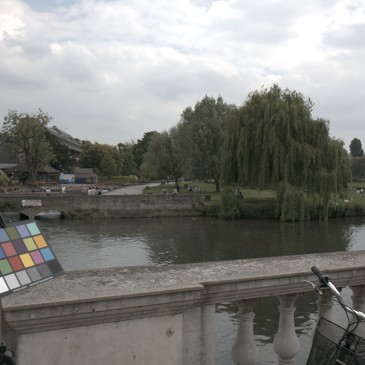}
    \hspace{-0.05in}
     \includegraphics[width=0.156in]{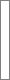}
    }
    {$\hat W$, $\hat L$, err = \input{figures/results/051_err_ours.txt}\textdegree}
    }
    \hspace{0.13in}
    \stackunder[5pt]{
     CM 19-Color \cite{Finlayson2013}
    }{
    \stackunder[5pt]{
    \includegraphics[width=1.25in]{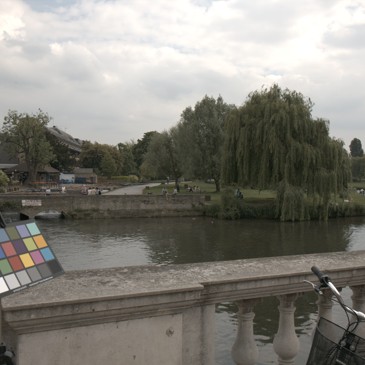}
    \hspace{-0.05in}
     \includegraphics[width=0.156in]{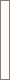}
     }
    { $\hat W$, $\hat L$, err = \input{figures/results/051_err_color.txt}\textdegree}
    }
    \hspace{0.13in}
    \stackunder[5pt]{
     CM 19-Edge \cite{Finlayson2013}
    }{
    \stackunder[5pt]{
    \includegraphics[width=1.25in]{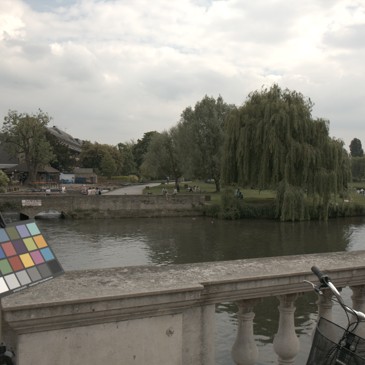}
    \hspace{-0.05in}
     \includegraphics[width=0.156in]{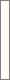}
     }
    { $\hat W$, $\hat L$, err = \input{figures/results/051_err_edge.txt}\textdegree}
    }
    \hspace{0.13in}
    \stackunder[5pt]{
     Cheng \etal, $p = 3.5$, \cite{Cheng14}
    }{
    \stackunder[5pt]{
    \includegraphics[width=1.25in]{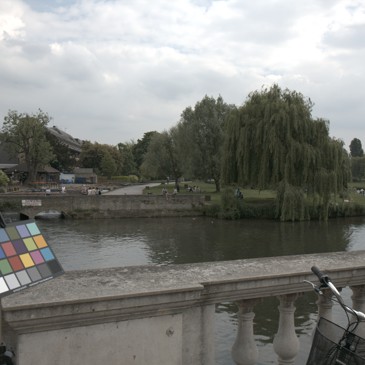}
    \hspace{-0.05in}
     \includegraphics[width=0.156in]{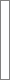}
     }
    { $\hat W$, $\hat L$, err = \input{figures/results/051_err_cheng.txt}\textdegree}
    }
    }
}
    \caption{ For each scene we present the input image produced by the camera alongside the ground-truth illumination color and white-balanced image. Images are shown in sRGB, normalized to the 98th percentile. For our algorithm and three baseline algorithms we show the estimated illumination and white-balanced image, as well as the error in degrees of the estimated illumination with respect to the ground-truth. Recovered illuminations are rendered with respect to ground-truth, such that ``white'' is correct, and any deviation from white is an error.
    \label{fig:problem1}
}
\end{figure*}

\begin{figure*}[b!]
\centering
  \stackunder[40pt]{
    \stackunder[10pt]{
    \stackunder[5pt]{Input image and ground-truth solution}
    {
    \stackunder[5pt]{
      \includegraphics[width=1.25in]{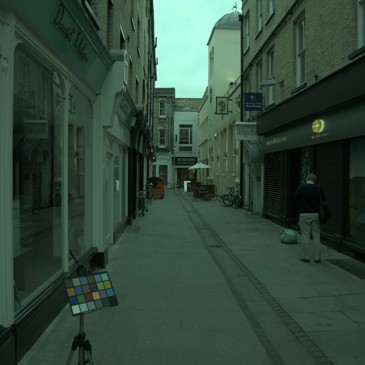}
      }{$I$}
    \stackunder[5pt]{
      \hspace{0.0in}
      }{$=$}
    \stackunder[5pt]{
      \includegraphics[width=1.25in]{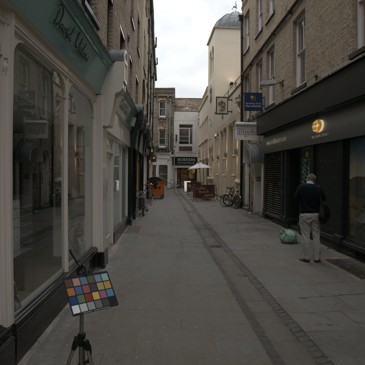}
    }{$W$}
    \stackunder[5pt]{
      \hspace{0.0in}
    }{$\times$}
    \stackunder[5pt]{
      \includegraphics[width=0.156in]{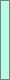}
      }{$L$}
    }}{
    \stackunder[5pt]{
     CCC
    }{
    \stackunder[5pt]{
    \includegraphics[width=1.25in]{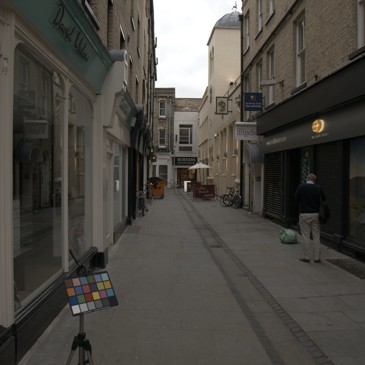}
    \hspace{-0.05in}
     \includegraphics[width=0.156in]{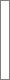}
    }
    {$\hat W$, $\hat L$, err = \input{figures/results/101_err_ours.txt}\textdegree}
    }
    \hspace{0.13in}
    \stackunder[5pt]{
     CM 19-Color \cite{Finlayson2013}
    }{
    \stackunder[5pt]{
    \includegraphics[width=1.25in]{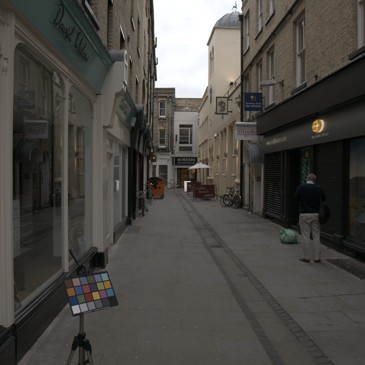}
    \hspace{-0.05in}
     \includegraphics[width=0.156in]{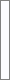}
     }
    { $\hat W$, $\hat L$, err = \input{figures/results/101_err_color.txt}\textdegree}
    }
    \hspace{0.13in}
    \stackunder[5pt]{
     CM 19-Edge \cite{Finlayson2013}
    }{
    \stackunder[5pt]{
    \includegraphics[width=1.25in]{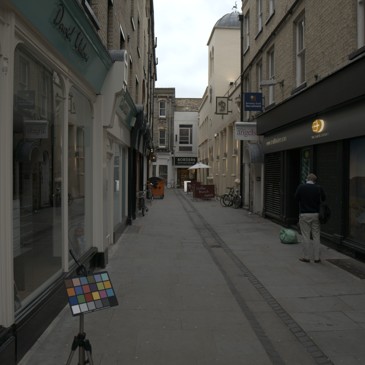}
    \hspace{-0.05in}
     \includegraphics[width=0.156in]{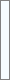}
     }
    { $\hat W$, $\hat L$, err = \input{figures/results/101_err_edge.txt}\textdegree}
    }
    \hspace{0.13in}
    \stackunder[5pt]{
     Cheng \etal, $p = 3.5$, \cite{Cheng14}
    }{
    \stackunder[5pt]{
    \includegraphics[width=1.25in]{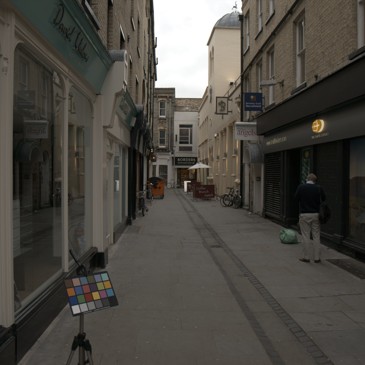}
    \hspace{-0.05in}
     \includegraphics[width=0.156in]{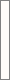}
     }
    { $\hat W$, $\hat L$, err = \input{figures/results/101_err_cheng.txt}\textdegree}
    }
    }
}{
    \stackunder[10pt]{
    \stackunder[5pt]{Input image and ground-truth solution}
    {
    \stackunder[5pt]{
      \includegraphics[width=1.25in]{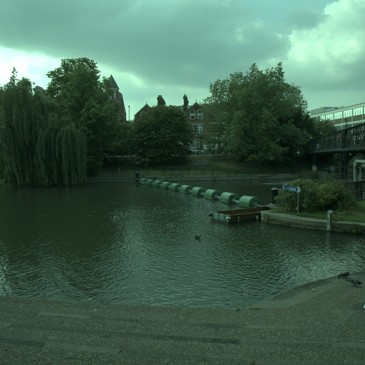}
      }{$I$}
    \stackunder[5pt]{
      \hspace{0.0in}
      }{$=$}
    \stackunder[5pt]{
      \includegraphics[width=1.25in]{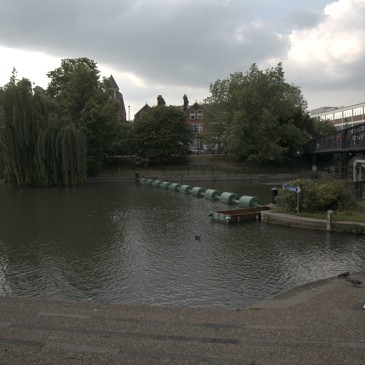}
    }{$W$}
    \stackunder[5pt]{
      \hspace{0.0in}
    }{$\times$}
    \stackunder[5pt]{
      \includegraphics[width=0.156in]{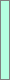}
      }{$L$}
    }}{
    \stackunder[5pt]{
     CCC
    }{
    \stackunder[5pt]{
    \includegraphics[width=1.25in]{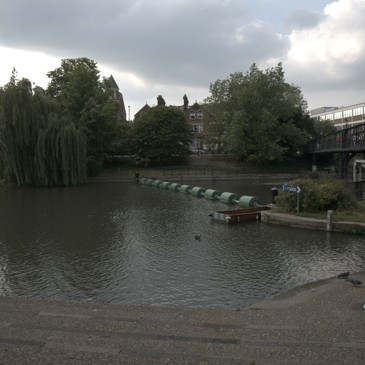}
    \hspace{-0.05in}
     \includegraphics[width=0.156in]{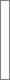}
    }
    {$\hat W$, $\hat L$, err = \input{figures/results/151_err_ours.txt}\textdegree}
    }
    \hspace{0.13in}
    \stackunder[5pt]{
     CM 19-Color \cite{Finlayson2013}
    }{
    \stackunder[5pt]{
    \includegraphics[width=1.25in]{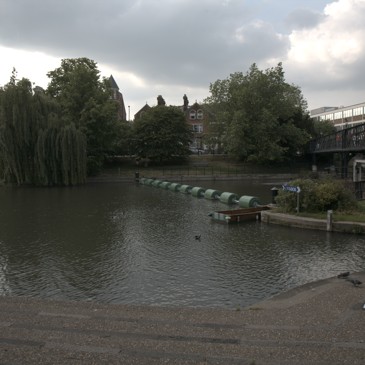}
    \hspace{-0.05in}
     \includegraphics[width=0.156in]{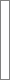}
     }
    { $\hat W$, $\hat L$, err = \input{figures/results/151_err_color.txt}\textdegree}
    }
    \hspace{0.13in}
    \stackunder[5pt]{
     CM 19-Edge \cite{Finlayson2013}
    }{
    \stackunder[5pt]{
    \includegraphics[width=1.25in]{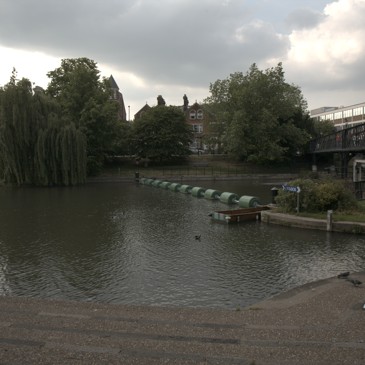}
    \hspace{-0.05in}
     \includegraphics[width=0.156in]{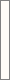}
     }
    { $\hat W$, $\hat L$, err = \input{figures/results/151_err_edge.txt}\textdegree}
    }
    \hspace{0.13in}
    \stackunder[5pt]{
     Cheng \etal, $p = 3.5$, \cite{Cheng14}
    }{
    \stackunder[5pt]{
    \includegraphics[width=1.25in]{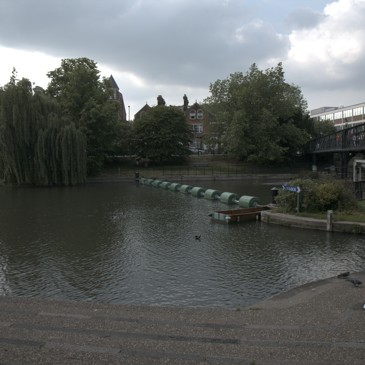}
    \hspace{-0.05in}
     \includegraphics[width=0.156in]{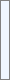}
     }
    { $\hat W$, $\hat L$, err = \input{figures/results/151_err_cheng.txt}\textdegree}
    }
    }
}
    \caption{Additional results in the same format as Figure~\ref{fig:problem1}.
}
\end{figure*}

\begin{figure*}[b!]
\centering
  \stackunder[40pt]{
    \stackunder[10pt]{
    \stackunder[5pt]{Input image and ground-truth solution}
    {
    \stackunder[5pt]{
      \includegraphics[width=1.25in]{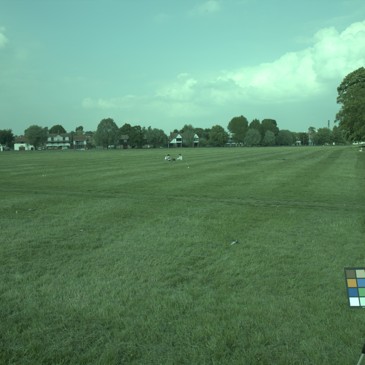}
      }{$I$}
    \stackunder[5pt]{
      \hspace{0.0in}
      }{$=$}
    \stackunder[5pt]{
      \includegraphics[width=1.25in]{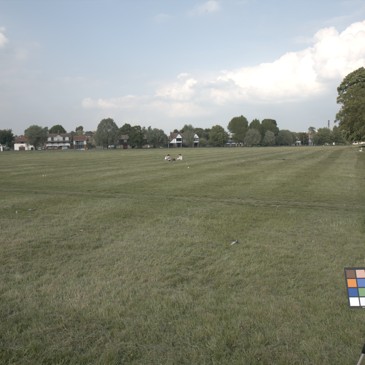}
    }{$W$}
    \stackunder[5pt]{
      \hspace{0.0in}
    }{$\times$}
    \stackunder[5pt]{
      \includegraphics[width=0.156in]{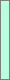}
      }{$L$}
    }}{
    \stackunder[5pt]{
     CCC
    }{
    \stackunder[5pt]{
    \includegraphics[width=1.25in]{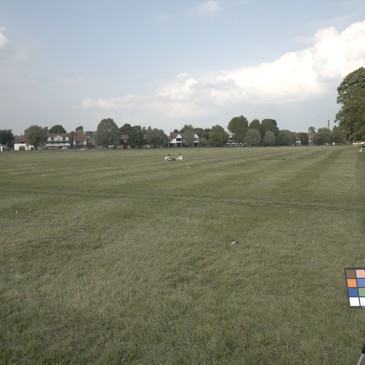}
    \hspace{-0.05in}
     \includegraphics[width=0.156in]{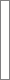}
    }
    {$\hat W$, $\hat L$, err = \input{figures/results/201_err_ours.txt}\textdegree}
    }
    \hspace{0.13in}
    \stackunder[5pt]{
     CM 19-Color \cite{Finlayson2013}
    }{
    \stackunder[5pt]{
    \includegraphics[width=1.25in]{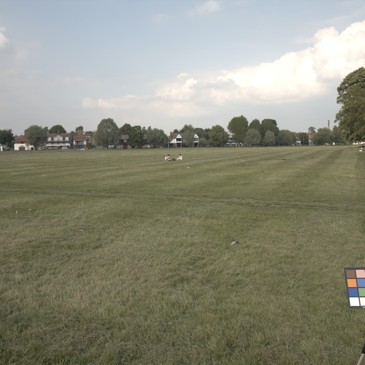}
    \hspace{-0.05in}
     \includegraphics[width=0.156in]{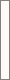}
     }
    { $\hat W$, $\hat L$, err = \input{figures/results/201_err_color.txt}\textdegree}
    }
    \hspace{0.13in}
    \stackunder[5pt]{
     CM 19-Edge \cite{Finlayson2013}
    }{
    \stackunder[5pt]{
    \includegraphics[width=1.25in]{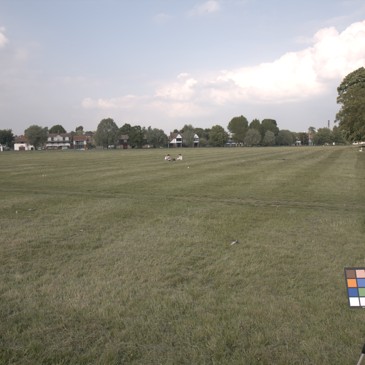}
    \hspace{-0.05in}
     \includegraphics[width=0.156in]{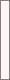}
     }
    { $\hat W$, $\hat L$, err = \input{figures/results/201_err_edge.txt}\textdegree}
    }
    \hspace{0.13in}
    \stackunder[5pt]{
     Cheng \etal, $p = 3.5$, \cite{Cheng14}
    }{
    \stackunder[5pt]{
    \includegraphics[width=1.25in]{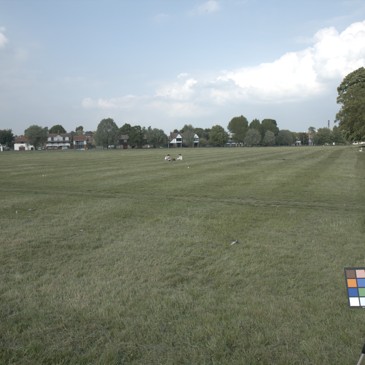}
    \hspace{-0.05in}
     \includegraphics[width=0.156in]{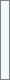}
     }
    { $\hat W$, $\hat L$, err = \input{figures/results/201_err_cheng.txt}\textdegree}
    }
    }
}{
    \stackunder[10pt]{
    \stackunder[5pt]{Input image and ground-truth solution}
    {
    \stackunder[5pt]{
      \includegraphics[width=1.25in]{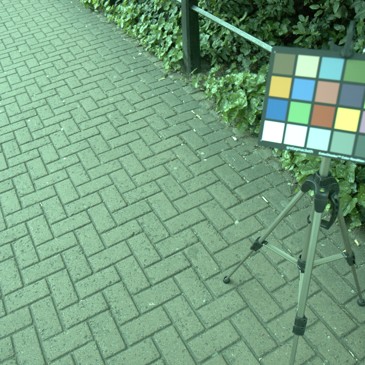}
      }{$I$}
    \stackunder[5pt]{
      \hspace{0.0in}
      }{$=$}
    \stackunder[5pt]{
      \includegraphics[width=1.25in]{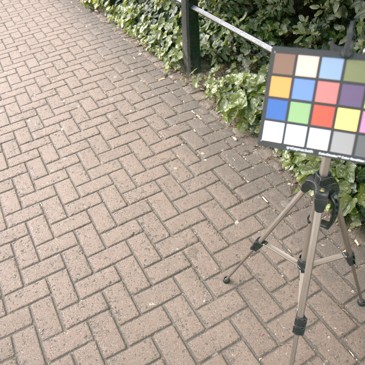}
    }{$W$}
    \stackunder[5pt]{
      \hspace{0.0in}
    }{$\times$}
    \stackunder[5pt]{
      \includegraphics[width=0.156in]{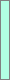}
      }{$L$}
    }}{
    \stackunder[5pt]{
     CCC
    }{
    \stackunder[5pt]{
    \includegraphics[width=1.25in]{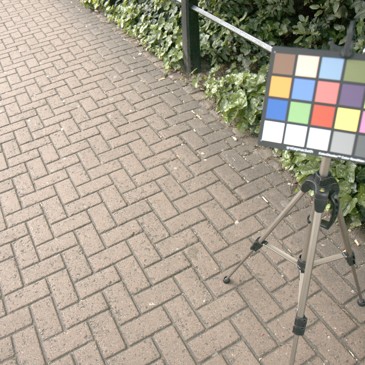}
    \hspace{-0.05in}
     \includegraphics[width=0.156in]{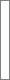}
    }
    {$\hat W$, $\hat L$, err = \input{figures/results/251_err_ours.txt}\textdegree}
    }
    \hspace{0.13in}
    \stackunder[5pt]{
     CM 19-Color \cite{Finlayson2013}
    }{
    \stackunder[5pt]{
    \includegraphics[width=1.25in]{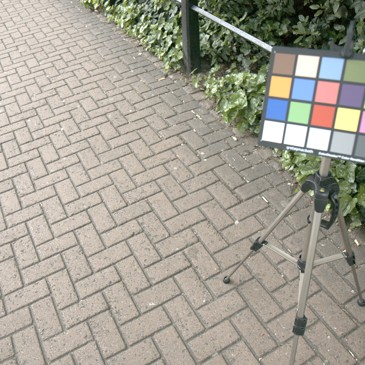}
    \hspace{-0.05in}
     \includegraphics[width=0.156in]{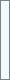}
     }
    { $\hat W$, $\hat L$, err = \input{figures/results/251_err_color.txt}\textdegree}
    }
    \hspace{0.13in}
    \stackunder[5pt]{
     CM 19-Edge \cite{Finlayson2013}
    }{
    \stackunder[5pt]{
    \includegraphics[width=1.25in]{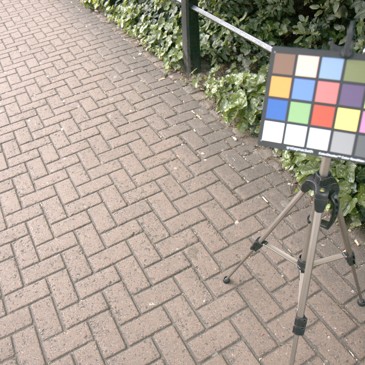}
    \hspace{-0.05in}
     \includegraphics[width=0.156in]{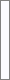}
     }
    { $\hat W$, $\hat L$, err = \input{figures/results/251_err_edge.txt}\textdegree}
    }
    \hspace{0.13in}
    \stackunder[5pt]{
     Cheng \etal, $p = 3.5$, \cite{Cheng14}
    }{
    \stackunder[5pt]{
    \includegraphics[width=1.25in]{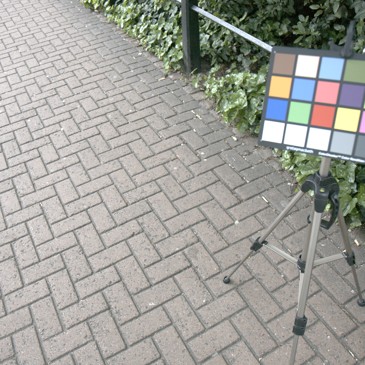}
    \hspace{-0.05in}
     \includegraphics[width=0.156in]{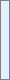}
     }
    { $\hat W$, $\hat L$, err = \input{figures/results/251_err_cheng.txt}\textdegree}
    }
    }
}
    \caption{Additional results in the same format as Figure~\ref{fig:problem1}.
}
\end{figure*}

\begin{figure*}[b!]
\centering
  \stackunder[40pt]{
    \stackunder[10pt]{
    \stackunder[5pt]{Input image and ground-truth solution}
    {
    \stackunder[5pt]{
      \includegraphics[width=1.25in]{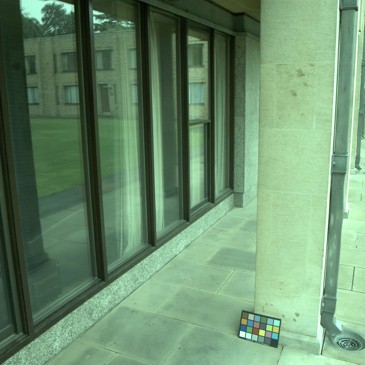}
      }{$I$}
    \stackunder[5pt]{
      \hspace{0.0in}
      }{$=$}
    \stackunder[5pt]{
      \includegraphics[width=1.25in]{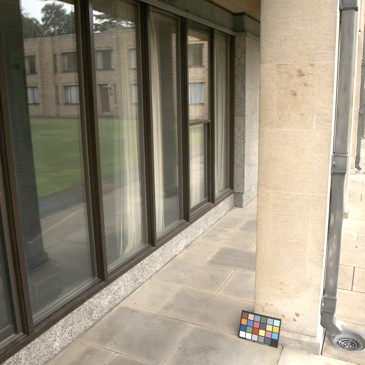}
    }{$W$}
    \stackunder[5pt]{
      \hspace{0.0in}
    }{$\times$}
    \stackunder[5pt]{
      \includegraphics[width=0.156in]{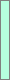}
      }{$L$}
    }}{
    \stackunder[5pt]{
     CCC
    }{
    \stackunder[5pt]{
    \includegraphics[width=1.25in]{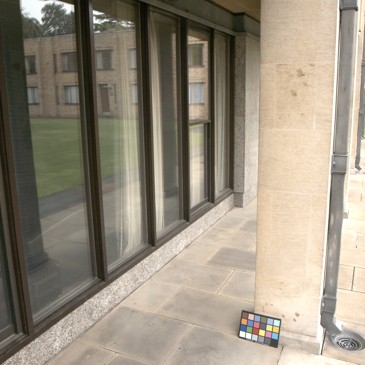}
    \hspace{-0.05in}
     \includegraphics[width=0.156in]{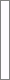}
    }
    {$\hat W$, $\hat L$, err = \input{figures/results/301_err_ours.txt}\textdegree}
    }
    \hspace{0.13in}
    \stackunder[5pt]{
     CM 19-Color \cite{Finlayson2013}
    }{
    \stackunder[5pt]{
    \includegraphics[width=1.25in]{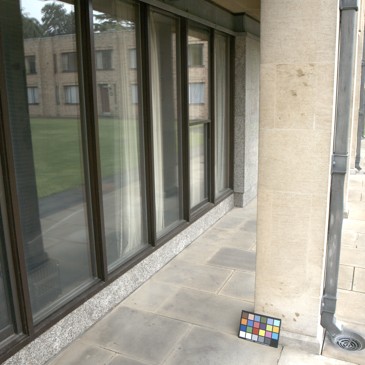}
    \hspace{-0.05in}
     \includegraphics[width=0.156in]{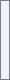}
     }
    { $\hat W$, $\hat L$, err = \input{figures/results/301_err_color.txt}\textdegree}
    }
    \hspace{0.13in}
    \stackunder[5pt]{
     CM 19-Edge \cite{Finlayson2013}
    }{
    \stackunder[5pt]{
    \includegraphics[width=1.25in]{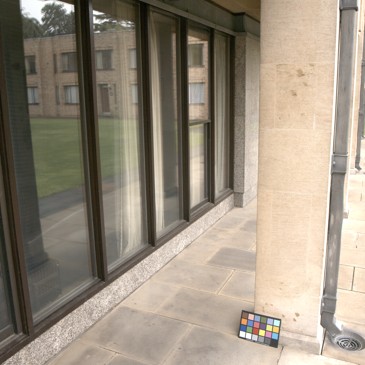}
    \hspace{-0.05in}
     \includegraphics[width=0.156in]{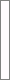}
     }
    { $\hat W$, $\hat L$, err = \input{figures/results/301_err_edge.txt}\textdegree}
    }
    \hspace{0.13in}
    \stackunder[5pt]{
     Cheng \etal, $p = 3.5$, \cite{Cheng14}
    }{
    \stackunder[5pt]{
    \includegraphics[width=1.25in]{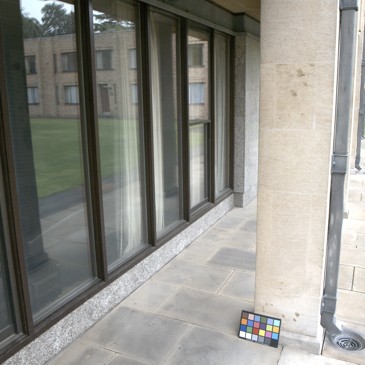}
    \hspace{-0.05in}
     \includegraphics[width=0.156in]{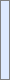}
     }
    { $\hat W$, $\hat L$, err = \input{figures/results/301_err_cheng.txt}\textdegree}
    }
    }
}{
    \stackunder[10pt]{
    \stackunder[5pt]{Input image and ground-truth solution}
    {
    \stackunder[5pt]{
      \includegraphics[width=1.25in]{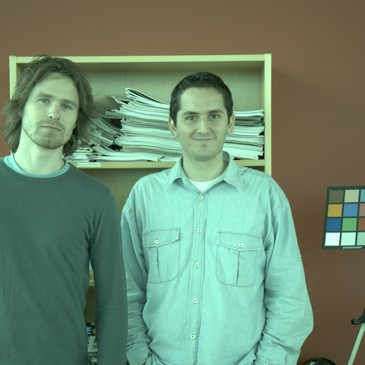}
      }{$I$}
    \stackunder[5pt]{
      \hspace{0.0in}
      }{$=$}
    \stackunder[5pt]{
      \includegraphics[width=1.25in]{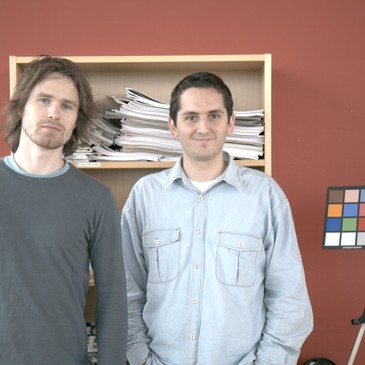}
    }{$W$}
    \stackunder[5pt]{
      \hspace{0.0in}
    }{$\times$}
    \stackunder[5pt]{
      \includegraphics[width=0.156in]{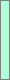}
      }{$L$}
    }}{
    \stackunder[5pt]{
     CCC
    }{
    \stackunder[5pt]{
    \includegraphics[width=1.25in]{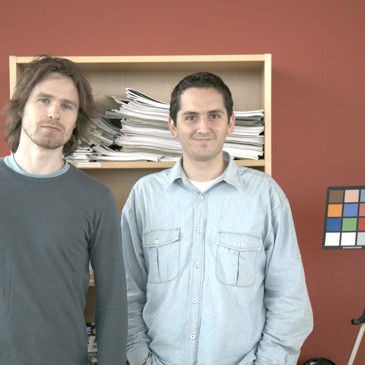}
    \hspace{-0.05in}
     \includegraphics[width=0.156in]{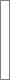}
    }
    {$\hat W$, $\hat L$, err = \input{figures/results/351_err_ours.txt}\textdegree}
    }
    \hspace{0.13in}
    \stackunder[5pt]{
     CM 19-Color \cite{Finlayson2013}
    }{
    \stackunder[5pt]{
    \includegraphics[width=1.25in]{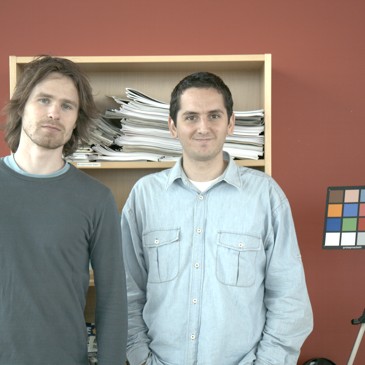}
    \hspace{-0.05in}
     \includegraphics[width=0.156in]{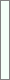}
     }
    { $\hat W$, $\hat L$, err = \input{figures/results/351_err_color.txt}\textdegree}
    }
    \hspace{0.13in}
    \stackunder[5pt]{
     CM 19-Edge \cite{Finlayson2013}
    }{
    \stackunder[5pt]{
    \includegraphics[width=1.25in]{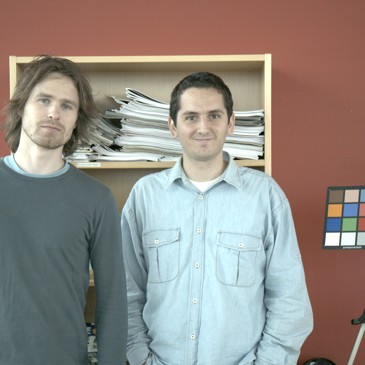}
    \hspace{-0.05in}
     \includegraphics[width=0.156in]{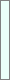}
     }
    { $\hat W$, $\hat L$, err = \input{figures/results/351_err_edge.txt}\textdegree}
    }
    \hspace{0.13in}
    \stackunder[5pt]{
     Cheng \etal, $p = 3.5$, \cite{Cheng14}
    }{
    \stackunder[5pt]{
    \includegraphics[width=1.25in]{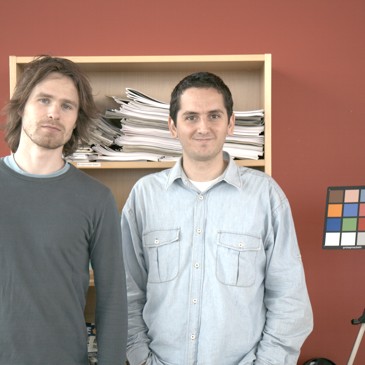}
    \hspace{-0.05in}
     \includegraphics[width=0.156in]{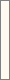}
     }
    { $\hat W$, $\hat L$, err = \input{figures/results/351_err_cheng.txt}\textdegree}
    }
    }
}
    \caption{Additional results in the same format as Figure~\ref{fig:problem1}.
}
\end{figure*}

\begin{figure*}[b!]
\centering
  \stackunder[40pt]{
    \stackunder[10pt]{
    \stackunder[5pt]{Input image and ground-truth solution}
    {
    \stackunder[5pt]{
      \includegraphics[width=1.25in]{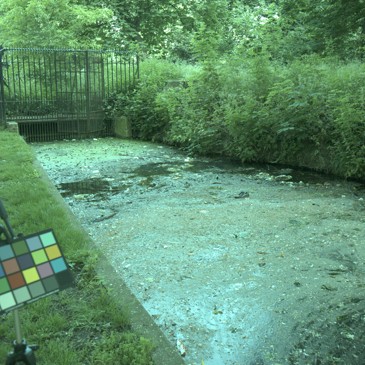}
      }{$I$}
    \stackunder[5pt]{
      \hspace{0.0in}
      }{$=$}
    \stackunder[5pt]{
      \includegraphics[width=1.25in]{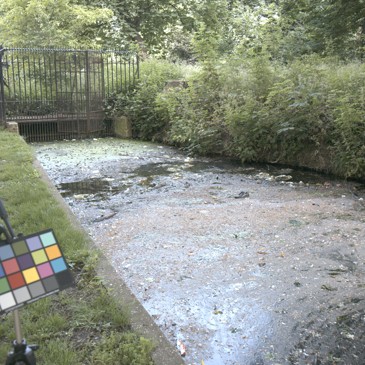}
    }{$W$}
    \stackunder[5pt]{
      \hspace{0.0in}
    }{$\times$}
    \stackunder[5pt]{
      \includegraphics[width=0.156in]{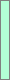}
      }{$L$}
    }}{
    \stackunder[5pt]{
     CCC
    }{
    \stackunder[5pt]{
    \includegraphics[width=1.25in]{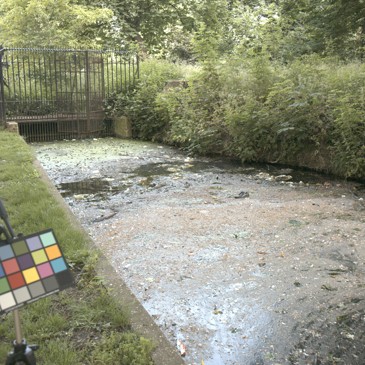}
    \hspace{-0.05in}
     \includegraphics[width=0.156in]{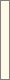}
    }
    {$\hat W$, $\hat L$, err = \input{figures/results/401_err_ours.txt}\textdegree}
    }
    \hspace{0.13in}
    \stackunder[5pt]{
     CM 19-Color \cite{Finlayson2013}
    }{
    \stackunder[5pt]{
    \includegraphics[width=1.25in]{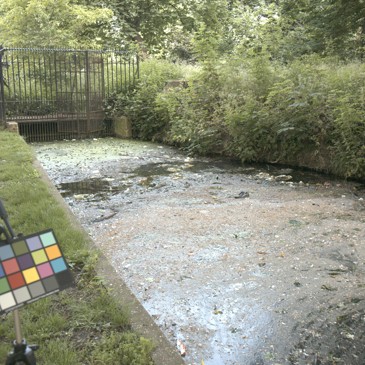}
    \hspace{-0.05in}
     \includegraphics[width=0.156in]{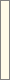}
     }
    { $\hat W$, $\hat L$, err = \input{figures/results/401_err_color.txt}\textdegree}
    }
    \hspace{0.13in}
    \stackunder[5pt]{
     CM 19-Edge \cite{Finlayson2013}
    }{
    \stackunder[5pt]{
    \includegraphics[width=1.25in]{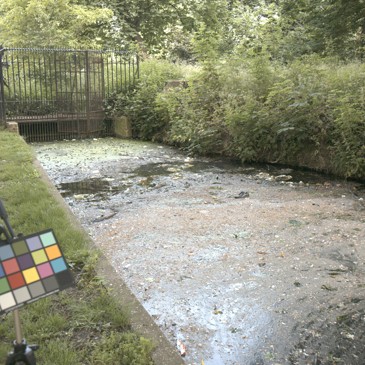}
    \hspace{-0.05in}
     \includegraphics[width=0.156in]{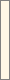}
     }
    { $\hat W$, $\hat L$, err = \input{figures/results/401_err_edge.txt}\textdegree}
    }
    \hspace{0.13in}
    \stackunder[5pt]{
     Cheng \etal, $p = 3.5$, \cite{Cheng14}
    }{
    \stackunder[5pt]{
    \includegraphics[width=1.25in]{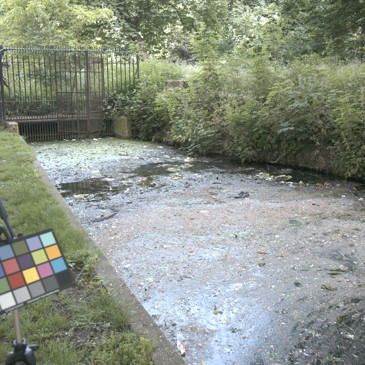}
    \hspace{-0.05in}
     \includegraphics[width=0.156in]{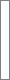}
     }
    { $\hat W$, $\hat L$, err = \input{figures/results/401_err_cheng.txt}\textdegree}
    }
    }
}{
    \stackunder[10pt]{
    \stackunder[5pt]{Input image and ground-truth solution}
    {
    \stackunder[5pt]{
      \includegraphics[width=1.25in]{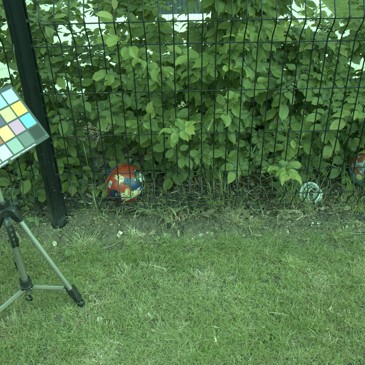}
      }{$I$}
    \stackunder[5pt]{
      \hspace{0.0in}
      }{$=$}
    \stackunder[5pt]{
      \includegraphics[width=1.25in]{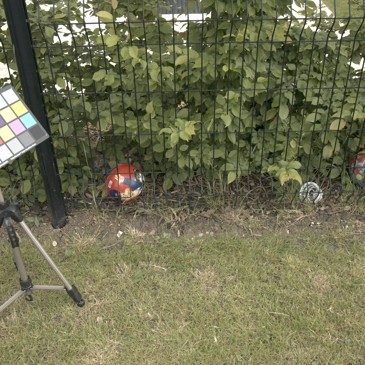}
    }{$W$}
    \stackunder[5pt]{
      \hspace{0.0in}
    }{$\times$}
    \stackunder[5pt]{
      \includegraphics[width=0.156in]{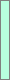}
      }{$L$}
    }}{
    \stackunder[5pt]{
     CCC
    }{
    \stackunder[5pt]{
    \includegraphics[width=1.25in]{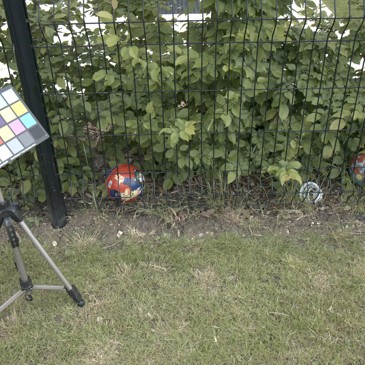}
    \hspace{-0.05in}
     \includegraphics[width=0.156in]{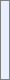}
    }
    {$\hat W$, $\hat L$, err = \input{figures/results/451_err_ours.txt}\textdegree}
    }
    \hspace{0.13in}
    \stackunder[5pt]{
     CM 19-Color \cite{Finlayson2013}
    }{
    \stackunder[5pt]{
    \includegraphics[width=1.25in]{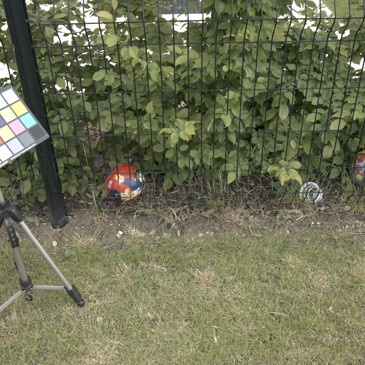}
    \hspace{-0.05in}
     \includegraphics[width=0.156in]{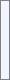}
     }
    { $\hat W$, $\hat L$, err = \input{figures/results/451_err_color.txt}\textdegree}
    }
    \hspace{0.13in}
    \stackunder[5pt]{
     CM 19-Edge \cite{Finlayson2013}
    }{
    \stackunder[5pt]{
    \includegraphics[width=1.25in]{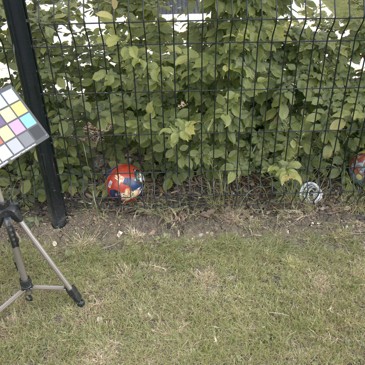}
    \hspace{-0.05in}
     \includegraphics[width=0.156in]{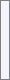}
     }
    { $\hat W$, $\hat L$, err = \input{figures/results/451_err_edge.txt}\textdegree}
    }
    \hspace{0.13in}
    \stackunder[5pt]{
     Cheng \etal, $p = 3.5$, \cite{Cheng14}
    }{
    \stackunder[5pt]{
    \includegraphics[width=1.25in]{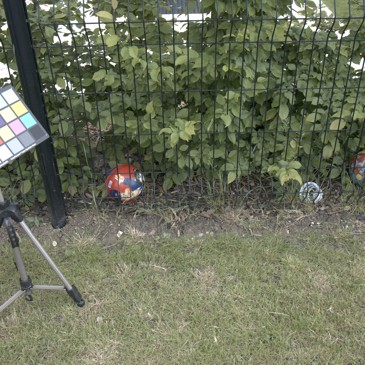}
    \hspace{-0.05in}
     \includegraphics[width=0.156in]{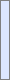}
     }
    { $\hat W$, $\hat L$, err = \input{figures/results/451_err_cheng.txt}\textdegree}
    }
    }
}
    \caption{Additional results in the same format as Figure~\ref{fig:problem1}.
}
\end{figure*}

\begin{figure*}[b!]
\centering
  \stackunder[40pt]{
    \stackunder[10pt]{
    \stackunder[5pt]{Input image and ground-truth solution}
    {
    \stackunder[5pt]{
      \includegraphics[width=1.25in]{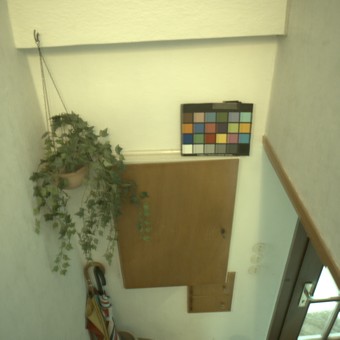}
      }{$I$}
    \stackunder[5pt]{
      \hspace{0.0in}
      }{$=$}
    \stackunder[5pt]{
      \includegraphics[width=1.25in]{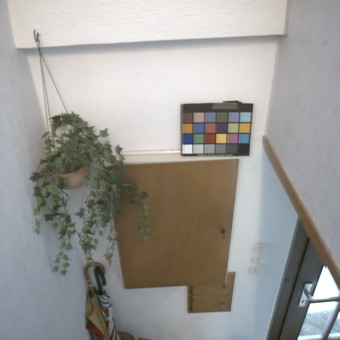}
    }{$W$}
    \stackunder[5pt]{
      \hspace{0.0in}
    }{$\times$}
    \stackunder[5pt]{
      \includegraphics[width=0.156in]{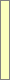}
      }{$L$}
    }}{
    \stackunder[5pt]{
     CCC
    }{
    \stackunder[5pt]{
    \includegraphics[width=1.25in]{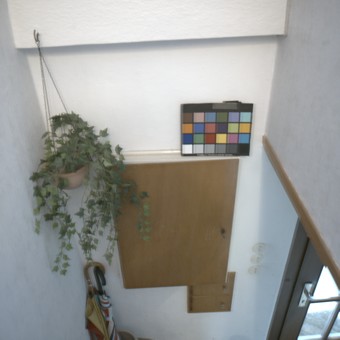}
    \hspace{-0.05in}
     \includegraphics[width=0.156in]{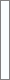}
    }
    {$\hat W$, $\hat L$, err = \input{figures/results/501_err_ours.txt}\textdegree}
    }
    \hspace{0.13in}
    \stackunder[5pt]{
     CM 19-Color \cite{Finlayson2013}
    }{
    \stackunder[5pt]{
    \includegraphics[width=1.25in]{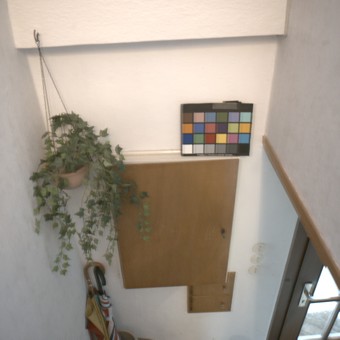}
    \hspace{-0.05in}
     \includegraphics[width=0.156in]{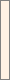}
     }
    { $\hat W$, $\hat L$, err = \input{figures/results/501_err_color.txt}\textdegree}
    }
    \hspace{0.13in}
    \stackunder[5pt]{
     CM 19-Edge \cite{Finlayson2013}
    }{
    \stackunder[5pt]{
    \includegraphics[width=1.25in]{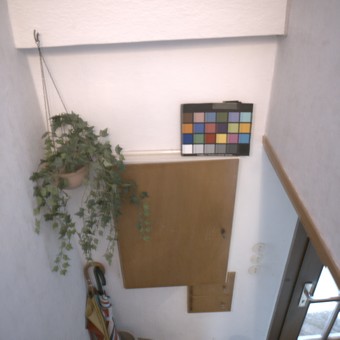}
    \hspace{-0.05in}
     \includegraphics[width=0.156in]{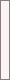}
     }
    { $\hat W$, $\hat L$, err = \input{figures/results/501_err_edge.txt}\textdegree}
    }
    \hspace{0.13in}
    \stackunder[5pt]{
     Cheng \etal, $p = 3.5$, \cite{Cheng14}
    }{
    \stackunder[5pt]{
    \includegraphics[width=1.25in]{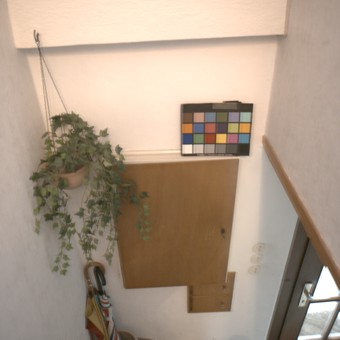}
    \hspace{-0.05in}
     \includegraphics[width=0.156in]{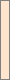}
     }
    { $\hat W$, $\hat L$, err = \input{figures/results/501_err_cheng.txt}\textdegree}
    }
    }
}{
    \stackunder[10pt]{
    \stackunder[5pt]{Input image and ground-truth solution}
    {
    \stackunder[5pt]{
      \includegraphics[width=1.25in]{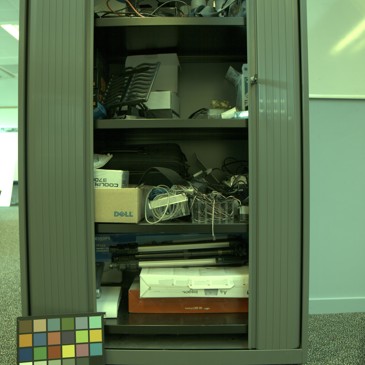}
      }{$I$}
    \stackunder[5pt]{
      \hspace{0.0in}
      }{$=$}
    \stackunder[5pt]{
      \includegraphics[width=1.25in]{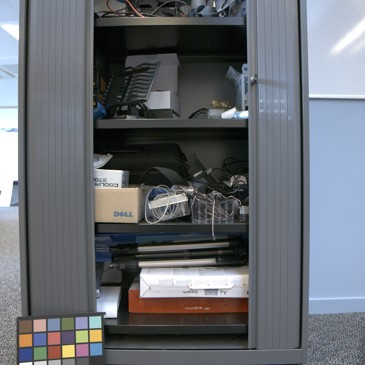}
    }{$W$}
    \stackunder[5pt]{
      \hspace{0.0in}
    }{$\times$}
    \stackunder[5pt]{
      \includegraphics[width=0.156in]{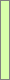}
      }{$L$}
    }}{
    \stackunder[5pt]{
     CCC
    }{
    \stackunder[5pt]{
    \includegraphics[width=1.25in]{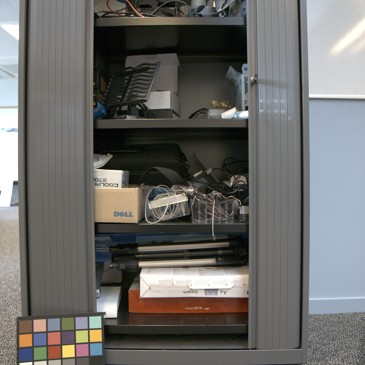}
    \hspace{-0.05in}
     \includegraphics[width=0.156in]{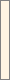}
    }
    {$\hat W$, $\hat L$, err = \input{figures/results/551_err_ours.txt}\textdegree}
    }
    \hspace{0.13in}
    \stackunder[5pt]{
     CM 19-Color \cite{Finlayson2013}
    }{
    \stackunder[5pt]{
    \includegraphics[width=1.25in]{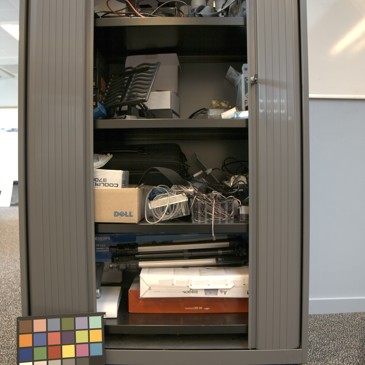}
    \hspace{-0.05in}
     \includegraphics[width=0.156in]{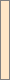}
     }
    { $\hat W$, $\hat L$, err = \input{figures/results/551_err_color.txt}\textdegree}
    }
    \hspace{0.13in}
    \stackunder[5pt]{
     CM 19-Edge \cite{Finlayson2013}
    }{
    \stackunder[5pt]{
    \includegraphics[width=1.25in]{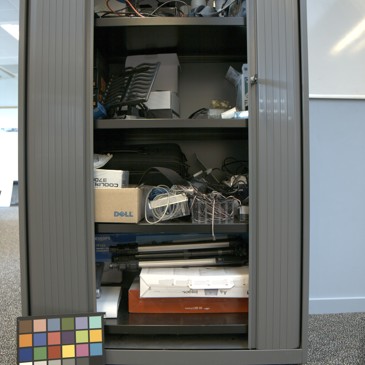}
    \hspace{-0.05in}
     \includegraphics[width=0.156in]{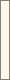}
     }
    { $\hat W$, $\hat L$, err = \input{figures/results/551_err_edge.txt}\textdegree}
    }
    \hspace{0.13in}
    \stackunder[5pt]{
     Cheng \etal, $p = 3.5$, \cite{Cheng14}
    }{
    \stackunder[5pt]{
    \includegraphics[width=1.25in]{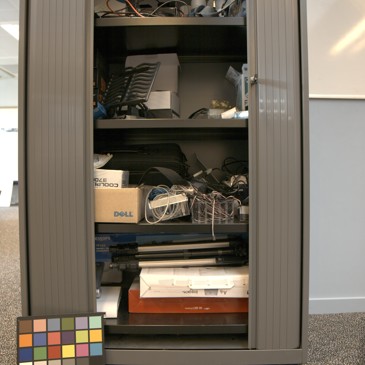}
    \hspace{-0.05in}
     \includegraphics[width=0.156in]{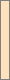}
     }
    { $\hat W$, $\hat L$, err = \input{figures/results/551_err_cheng.txt}\textdegree}
    }
    }
}
    \caption{Additional results in the same format as Figure~\ref{fig:problem1}.
}
\end{figure*}

{\small
\bibliographystyle{ieee}
\bibliography{white_balance}
}